\documentclass[10pt,twocolumn,letterpaper]{article} 
\usepackage{simpleConference}
\usepackage{times}
\usepackage{graphicx}
\usepackage{amssymb}
\usepackage{url,hyperref}
\usepackage{amsmath}

\usepackage{pifont}
\newcommand{\cmark}{\ding{51}}%
\newcommand{\xmark}{\ding{55}}%

\usepackage{authblk}
\usepackage{colortbl}
\usepackage{booktabs}
\usepackage{multirow}
\usepackage{tablefootnote}
\usepackage[dvipsnames]{xcolor}
\usepackage{color}
\usepackage[labelsep=space]{subcaption}
\usepackage{enumitem}
\usepackage[ruled,vlined]{algorithm2e}
\usepackage{algorithmic}
\definecolor{commentcolor}{RGB}{110,154,155}   
\newcommand{\PyComment}[1]{\ttfamily\textcolor{commentcolor}{\# #1}}  
\newcommand{\PyCode}[1]{\ttfamily\textcolor{black}{#1}} 

\usepackage[capitalize]{cleveref}
\crefname{section}{Sec.}{Secs.}
\Crefname{section}{Section}{Sections}
\Crefname{table}{Table}{Tables}
\crefname{table}{Tab.}{Tabs.}
\crefname{figure}{Figure}{Figures}

\usepackage{xspace}

\begin{document}

\title{ESceme: Vision-and-Language Navigation with Episodic Scene Memory}

\author[1,2]{Qi Zheng}
\author[3]{Daqing Liu}
\author[3]{Chaoyue Wang}
\author[2]{Jing Zhang}
\author[4]{Dadong Wang}
\author[2]{Dacheng Tao}
\affil[1]{Shenzhen University}
\affil[2]{University of Sydney}
\affil[3]{JD Explore Academy}
\affil[4]{DATA61, CSIRO}

\maketitle

\vspace{-0.5cm}
\begin{abstract}
\textit{\small
Vision-and-language navigation (VLN) simulates a visual agent that follows natural-language navigation instructions in real-world scenes. Existing approaches have made enormous progress in navigation in new environments, such as beam search, pre-exploration, and dynamic or hierarchical history encoding. To balance generalization and efficiency, we resort to memorizing visited scenarios apart from the ongoing route while navigating. In this work, we introduce a mechanism of Episodic Scene memory (ESceme) for VLN that wakes an agent's memories of past visits when it enters the current scene. The episodic scene memory allows the agent to envision a bigger picture of the next prediction. This way, the agent learns to utilize dynamically updated information instead of merely adapting to the current observations. We provide a simple yet effective implementation of ESceme by enhancing the accessible views at each location and progressively completing the memory while navigating. We verify the superiority of ESceme on short-horizon (R2R), long-horizon (R4R), and vision-and-dialog (CVDN) VLN tasks. Our ESceme also wins first place on the CVDN leaderboard. Code is available: \url{https://github.com/qizhust/esceme}.}
\end{abstract}

\section{Introduction} \label{sec:intr}

With breakthroughs in computer vision and natural language understanding, the embodiment hypothesis that an intelligent agent is born from its interaction with environments~\cite{smith2005development} is now attracting more and more attention to embodied AI tasks such as vision-and-language navigation (VLN). VLN is firstly defined in~\cite{anderson2018vision} towards the goal of a robot carrying out general verbal instructions, where an agent is required to follow natural-language instructions based on what it sees and adapt to previously unseen environments. VLN has developed various settings, such as fine-grained and short-horizon navigation (e.g., R2R~\cite{anderson2018vision} and RxR~\cite{ku2020room}), long-horizon navigation (e.g., R4R~\cite{jain2019stay}), vision-and-dialogue navigation (e.g., CVDN~\cite{thomason2020vision}), and navigation with high-level instructions (e.g., REVERIE~\cite{qi2020reverie}). Compared with non-embodied VL tasks such as visual question answering~\cite{antol2015vqa} and visual captioning~\cite{chen2015microsoft,xu2016msr}, VLN agents suffer from domain shifts and changing observations during multi-step decision-making in the scenarios.

\begin{figure}[th]
    \centering
    \begin{subfigure}[b]{0.47\textwidth}
         \centering
         \includegraphics[width=\textwidth]{./demo_crop.pdf}
     \end{subfigure}
     \hfill
     \begin{subfigure}[b]{0.47\textwidth}
         \centering
         \includegraphics[width=\textwidth]{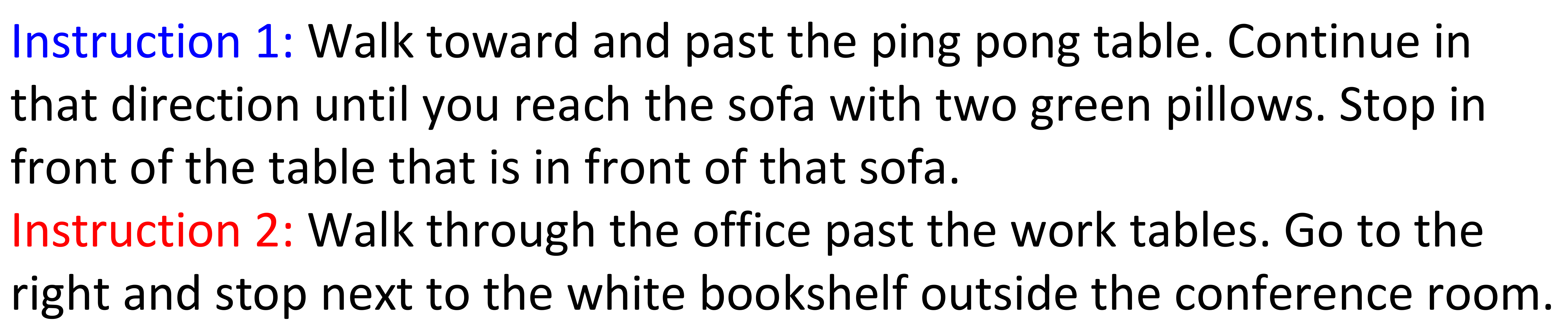}
     \end{subfigure}
    \caption{The \textcolor{blue}{blue} trajectory shows an agent carrying out instruction 1. The next time, the agent enters this scene to conduct the second instruction along the \textcolor{red}{red} path. ESceme allows it to recall the visited nodes (i.e., the \textcolor{blue}{blue} ones) at where it is standing (A) and choose the neighboring node B$_1$ that will see “the white bookshelf” in one more step at C. Finally, it navigates towards the \textcolor{red}{red} dash route and reaches the target.}
    \label{fig:demo}
\end{figure}

A vanilla Seq2Seq pipeline~\cite{anderson2018vision} that implicitly encodes path history with LSTMs~\cite{hochreiter1997long} shows moderate navigating ability. Since then, VLN performance has been considerably improved by pre-training~\cite{hao2020towards,hong2021vln,chen2021history,qiao2022hop}, data augmentations~\cite{fried2018speaker,tan2019learning,li2022envedit}, and algorithms that explicitly track past decisions along the trajectory~\cite{chen2021history,wang2021structured,chen2022think}. These methods learn enhanced representations by training VLN agents in each episode but ignore the dynamics of navigating over the whole data. Different strategies, including modified beam search~\cite{fried2018speaker} and pre-exploration~\cite{wang2019reinforced,tan2019learning,majumdar2020improving,zhu2020vision}, are devised to specifically increase adaptation to unseen environments at the cost of efficiency. Specifically, beam search significantly extends route length and involves much more interactions with the environment; pre-exploration takes extra steps to gather information and train the agent with auxiliary objectives before it can conduct given instructions. Such strategies incur burdensome time and computational expenses in practical usage.

In this work, we propose a navigation mechanism with Episodic Scene memory (ESceme) to balance generalization and efficiency by exploiting the dynamics of navigating all the episodes. ESceme requires no extra annotations or heavy computation and is agent-agnostic. We encode observation, instruction, and path history separately and update the scene memory during navigation via candidate enhancing. By preserving the memory among episodes, ESceme envisions the agent seeing a bigger picture in each decision. This way, the agent learns to utilize dynamically updated information instead of merely adapting to the current observations. Then during inference, it predicts actions with the progressively completed memory. A demonstration is shown in Fig.~\ref{fig:demo}. When carrying out an instruction at Location A, the agent is to select one from the adjacent nodes B$_1$-B$_5$ to navigate. It recalls the episodic scene memory, i.e., the blue route of a completed trajectory, and chooses Node B$_1$ that will see \textit{``the white bookshelf''} in one more step at C. 

We verify the superiority of ESceme in short-horizon navigation with fine-grained instruction (R2R), long-horizon navigation (R4R), and vision-and-dialog navigation (CVDN). We find that ESceme notably benefits navigation with longer routes (R4R and CVDN), promoting both successful reaching and path fidelity. Our method achieves the highest Goal Progress in the CVDN challenge. Besides a fair comparison with existing approaches under a single run, we test the performance with an approximately complete memory, where the agent fully updates its scene memory in the first round of navigation over all the episodes. We denote it as ESceme*, which serves as the upper bound of ESceme. We observe a further improvement in ESceme*, which indicates better-completed memory magnifies the advantage of ESceme. We hope this work can inspire further explorations in modeling episodic scene memory for VLN.

Since ESceme does not introduce any extra time or steps before following the instruction in inference, it is fair to compare it with its counterparts in the single-run setting. Very different from pre-exploration optimizing the parameters of an agent before solving the task, ESceme only renews its episodic memory while conducting instructions and requires no back-propagation operations. Moreover, ESceme neither involves beam search nor changes the local action space in sequential decision-making. These properties make ESceme both efficient and effective in reality use.
Our contributions are summarized as follows:
\begin{itemize}[leftmargin=*]
    \item We devise the first navigation mechanism with episodic scene memory (ESceme) for VLN to balance generalization and efficiency.
    \item We provide a simple yet effective implementation of ESceme via candidate enhancing, tested with two navigation architectures and two inferring strategies.
    \item We verify the superiority of ESceme in short-horizon (R2R), long-horizon (R4R), and vision-and-dialog (CVDN) navigation, and achieve a new state-of-the-art.
\end{itemize}

\section{Related work} \label{sec:rel_w}
\subsection{Vision-and-language navigation}

Since~\cite{anderson2018vision} defined the VLN task and provided an LSTM-based sequence-to-sequence baseline (Seq2Seq), numerous approaches have been developed. A branch of methods improves navigation via data augmentation, such as SF~\cite{fried2018speaker}, EnvDrop~\cite{tan2019learning}, and EnvEdit~\cite{li2022envedit}. As for agent training,~\cite{wang2018look} model the environment to provide planned-ahead information during navigation. RCM~\cite{wang2019reinforced} provides an intrinsic reward for reinforcement learning via an instruction-trajectory matching critic.~\cite{wang2020environment} jointly train an agent on VLN and vision-dialog navigation (MT-RCM+EnvAg). To fully use available semantic information in the environment, AuxRN~\cite{zhu2020vision} devises four self-supervised auxiliary reasoning tasks. TDSTP~\cite{zhao2022target} introduces an extra target location estimation during finetuning to achieve reliable path planning. Many methods explore more effective feature representations and architectures, such as PTA~\cite{cornia2019perceive}, OAAM~\cite{qi2020object}, NvEM~\cite{an2021neighbor}, RelGraph~\cite{hong2020language}, MTVM~\cite{lin2021multimodal}, and SEvol~\cite{chen2022reinforced}. 

Some methods construct and reason about a graph of navigation while conducting an episode, such as NTS~\cite{chaplot2020neural} and RECON~\cite{shah2022rapid} in the ImageGoal space and ETPNav~\cite{an2023etpnav} and CMTP~\cite{chen2021topological} in the VLN space. VLN-SIG~\cite{li2023improving} adds the tasks of generating semantics for future navigation views in pre-training and fine-tuning, and contributes to a more powerful agent backbone. KERM~\cite{li2023kerm} introduces knowledge described by text to aid action prediction, which is useful mainly in seen environments. GridMM~\cite{wang2023gridmm} builds a grid memory with fine-grained features and adopts a global action space, which improves the success rate but suffers from a much longer trajectory length.

Inspired by the breakthrough of large-scale pre-trained BERT~\cite{kenton2019bert} in natural language processing tasks, PRESS~\cite{li2019robust} replaces RNNs with pre-trained BERT to encode instructions and achieves a non-trivial improvement in unseen environments. PREVELENT~\cite{hao2020towards} pre-trains BERT from scratch using image-text-action triplets and further boosts the performance. RecBERT~\cite{hong2021vln} integrates a recurrent unit into a BERT model to be time-aware. \cite{chen2021history} propose the first VLN network that allows a sequence of historical memory and can be optimized end-to-end (HAMT). HOP~\cite{qiao2022hop} designs trajectory order modeling and group order modeling tasks to model temporal order information in pre-training. CSAP~\cite{wu2022cross} proposes trajectory-conditioned masked fragment modeling and contrastive semantic-alignment modeling tasks for pre-training. ADAPT~\cite{lin2022adapt} explicitly learns action-level modality alignment with action prompts. There are also some works specially designed for vision-and-dialog navigation, such as VISITRON~\cite{shrivastava2022visitron}, SCoA~\cite{zhu2021self}, and CMN~\cite{zhu2020visionb}.

The differences between the proposed ESceme and previous graph- or map-construction approaches are twofold. First, they construct path-level memory along a route in a single conduction. ESceme maintains scene-level memory from multiple episodes in the same scenario. Second, they use the path-level memory in planning by extending the agent's action space from local to global. ESceme does not change the agent's action space. Instead, it improves navigation by increasing the information of each node, which is the core idea that makes ESceme perform better than path-level memory methods (e.g., EGP and SSM). Scene memory is also studied in other works~\cite{datta2022episodic,georgakis2022cross,li2022layout,krantz2023iterative,vasudevan2021talk2nav,gupta2017cognitive}.

The method that is the most closely related to ours is IVLN~\cite{krantz2023iterative}. Our setting is actually identical to IVLN, which reorganizes episodes into tours. We store scene IDs during inference instead of explicitly organizing episodes according to their IDs, yet the two ways result in the same effect. Although both explore the impact of episodic memory and compare with the same baseline model, i.e., HAMT, our work provides a more effective design of the memory mechanism and obtains better performance due to candidate enhancing. Instead, the episodic memory in IVLN encodes the memory map as a whole and observes even worse results than the path-level memory baseline (cf. Table 2 in IVLN).

\begin{figure*}[t]
    \centering
    \includegraphics[width=0.96\textwidth]{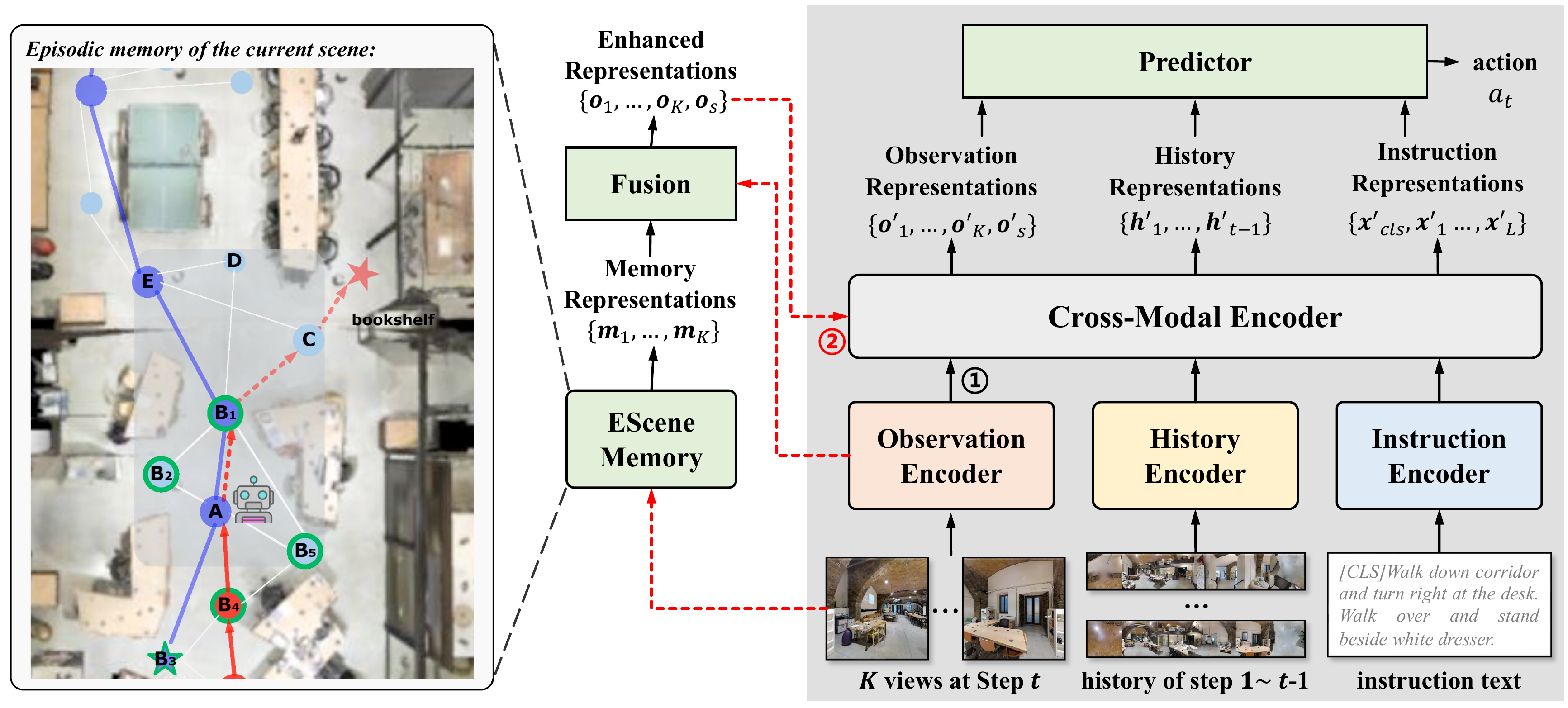}
    \caption{An overview of the \textbf{E}pisodic \textbf{Sce}ne \textbf{me}mory mechanism for VLN. On the left is partial episodic memory for the current scene, which gets updated in navigation 1) following the previous instruction, i.e., the \textcolor{blue}{blue} route, and 2) following the current instruction from Step 1 to $t-1$, i.e., the solid \textcolor{red}{red} trajectory. The \textcolor{cyan}{cyan} nodes are those viewed but not visited. The shadow box shows the memory of node B$_1$, which has six adjacent neighbors, i.e., A, B$_2$, B$_5$, C, D, and E. The integration of these nodes consists of the memory of B$_1$. At Step $t$, the agent stands at Node A and is expected to choose one node from B$_1$ to B$_5$. Given observation from K views, each view retrieves its memory in ESceme and produces $\{\mathbf{m}_1,...,\mathbf{m}_K\}$. The memory representation then fuses with original encoded observations, which yields $\{\mathbf{o}_1,...,\mathbf{o}_K,\mathbf{o}_s\}$. $o_s$ is the representation for STOP. The enhanced observations, instruction text, and history from Step 1 to $t-1$ compose the input to a navigation network to predict the action $a_t=i\in \{1,...,K,s\}$. Generally, a navigation network uses the encoded features of the original K views as the input to the cross-modal encoder, i.e., the output \textcircled{1}. Our ESceme exploits the enhanced observations from \textcircled{2}.}
    \label{fig:framework}
\end{figure*}

\subsection{Exploration strategies in VLN} As the navigation graph is pre-defined in discrete VLN, diverse strategies are adopted other than the regularly used single-run. For example,~\cite{fried2018speaker} modifies the standard beam search to select the final navigation route, which notably increases navigation success at the cost of unbearable trajectory lengths. More efficient pre-exploration methods are studied. For instance, a progress monitor is trained to discard unfinished trajectories during inference~\cite{ma2019self}. \cite{ma2019regretful} learn a regret module to decide when to backtrack.~\cite{ke2019tactical} compare partial paths with global information considered and backtrack only when necessary. AcPercep~\cite{wang2020active} learns an exploration policy to gather visual information for navigation. Although these methods improve searching efficiency, they heavily depend on manually designed or heuristic rules.~\cite{deng2020evolving} define a global action space for the first time and build a graphical representation of the environment for elegant exploration/backtracking.~\cite{wang2021structured} extend EnvDrop~\cite{tan2019learning} with an external structured scene memory (SSM) to promote exploration in the global action space. 

Pre-exploration, which allows an agent to pre-explore unseen environments before navigating, is first introduced in~\cite{wang2019reinforced} as a setting different from single-run and beam search. The obtained information functions in diverse ways. RCM~\cite{wang2019reinforced} uses the exploration experience in self-supervised imitation learning. EnvDrop~\cite{tan2019learning} exploits the environment information for data augmentation via back-translation. VLN-BERT~\cite{majumdar2020improving} provides the agent with a global view for optimal route selection. AuxRN~\cite{zhu2020vision} finetunes the agent in unseen environments with auxiliary tasks.

\begin{figure*}
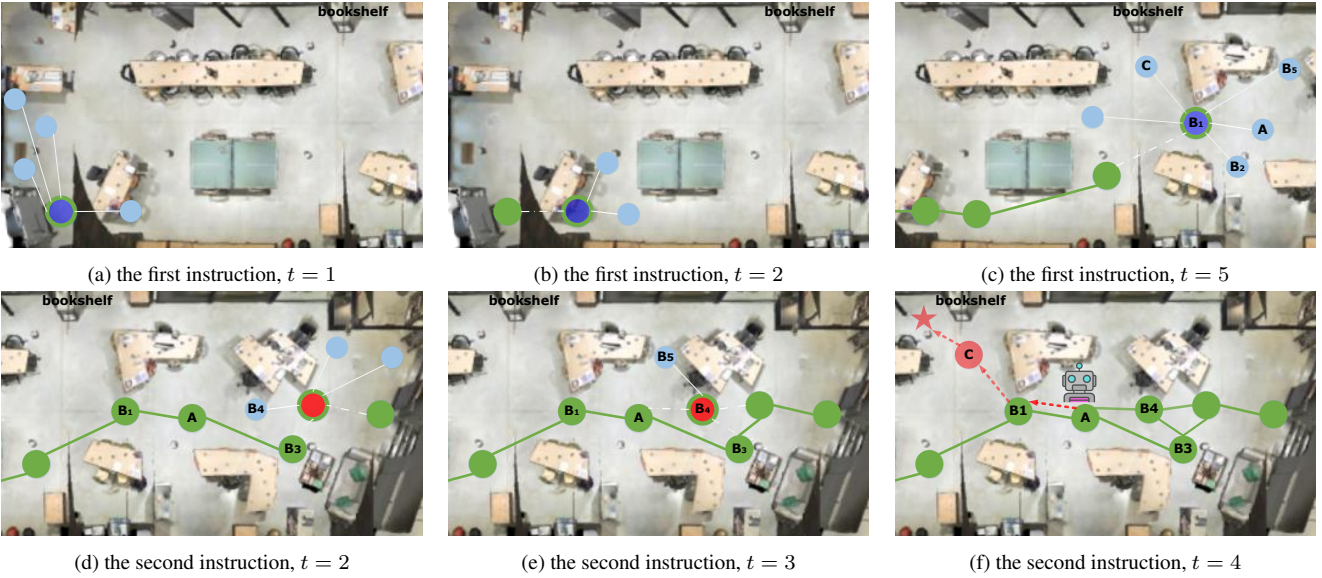

     \centering
     \begin{subfigure}[b]{0.32\textwidth}
         \centering
         \includegraphics[width=\textwidth]{./1t1.pdf}
         \caption{the first instruction, $t=1$}
         \label{fig:1t1}
     \end{subfigure}
     \hfill
     \begin{subfigure}[b]{0.32\textwidth}
         \centering
         \includegraphics[width=\textwidth]{./1t2.pdf}
         \caption{the first instruction, $t=2$}
         \label{fig:1t2}
     \end{subfigure}
     \hfill
     \begin{subfigure}[b]{0.32\textwidth}
         \centering
         \includegraphics[width=\textwidth]{./1t5.pdf}
         \caption{the first instruction, $t=5$}
         \label{fig:1t5}
     \end{subfigure}
     \hfill
     \begin{subfigure}[b]{0.32\textwidth}
         \centering
         \includegraphics[width=\textwidth]{./2t2.pdf}
         \caption{the second instruction, $t=2$}
         \label{fig:2t2}
     \end{subfigure}
     \hfill
     \begin{subfigure}[b]{0.32\textwidth}
         \centering
         \includegraphics[width=\textwidth]{./2t3.pdf}
         \caption{the second instruction, $t=3$}
         \label{fig:2t3}
     \end{subfigure}
     \hfill
     \begin{subfigure}[b]{0.32\textwidth}
         \centering
         \includegraphics[width=\textwidth]{./2t4.pdf}
         \caption{the second instruction, $t=4$}
         \label{fig:2t4}
     \end{subfigure}
        \caption{Episodic memory construction of a scene during navigation. ESceme at the beginning of each time step is presented in the figures, which comprises \textcolor{LimeGreen}{green} nodes and edges and is empty at the beginning of $t=1$. The \textcolor{blue}{blue} nodes indicate the current location of following the first instruction at each time step, and the \textcolor{red}{red} ones correspond to the second instruction. The small \textcolor{cyan}{cyan} nodes mark the remaining navigable viewpoints of the current location. Nodes with \textcolor{LimeGreen}{green} boundary are the chosen viewpoints in each time step. ESceme at the end of that time step is updated by the node with \textcolor{LimeGreen}{green} boundary and the dashed lines connected to its existing nodes. Please refer to Fig.~\ref{fig:demo} for a complete global graph of the scene, which is unavailable to the agent either in navigation or ESceme construction.}
        \label{fig:memory}
\end{figure*}

\section{Method} \label{sec:method}

\subsection{Problem formulation}

Given an instruction $X_i$, e.g., \textit{``Turn around and walk to the right of the room...''}, an agent starts from the initial location of route $R_i$. It observes a panoramic view of the environment $Y_i$. The panoramic view consists of $K{=}36$ single viewpoints, each of which is accompanied by an orientation $(\theta,\phi)$ indicating heading and elevation and a binary navigable signal. The agent selects a viewpoint from the navigable ones and moves to the next location with new observations. This process repeats until the agent takes the STOP action.

In a regular VLN task, there is a set of training samples $\mathcal{D}=\{(Y_1,X_1,R_1),...,(Y_{N_1},X_{N_1},R_{N_1})\}$, where $(X_i,R_i)$ is the instruction-route pair in an environment $Y_i$. The set $\{Y_1,...,Y_{N_1}\}$ denotes the seen environments during training. An agent is expected to learn navigation with $\mathcal{D}$ and carry out instructions in unseen scenarios given by $\mathcal{D}^u=\{(Y^u_1,X_1),...,(Y^u_{N_2},X_{N_2})\}$. The set $\{Y^u_1,...,Y^u_{N_2}\}$ denotes the unseen environments for test.

For a sequence prediction problem, history is an important source of information apart from observations and instructions. The shadow part in Fig.~\ref{fig:framework} shows a decision step by a general navigation approach that follows the pretraining-finetuning branch and encodes path history, represented by HAMT\cite{chen2021history}. We denote the vanilla features of $K$ single views extracted by the observation encoder as $\{\mathbf{f}_1,...,\mathbf{f}_K,\mathbf{f}_s\}$, which can be obtained by concatenating the separate features of encoded RGB images and orientations. $\mathbf{f}_s$ is appended to allow a STOP action. Together with history representations $\{\mathbf{h}_1,...,\mathbf{h}_{t-1}\}$ from the history encoder and text representations $\{\mathbf{x}_{cls},\mathbf{x}_1,...,$ $\mathbf{x}_L\}$ from the instruction encoder, the features of the observations \textcircled{1} are input into a cross-modal encoder for multi-modal fusion. A predictor block takes in the cross-modal representations $\{\mathbf{o}'_1,...,\mathbf{o}'_K,\mathbf{o}'_s\}$, $\{\mathbf{h}'_1,...,\mathbf{h}'_{t-1}\}$, and $\{\mathbf{x}'_{cls},\mathbf{x}'_1,...,\mathbf{x}'_L\}$ to predict action $a_t$.

Due to potential differences between seen and unseen environments, such as the appearance and layout of the scenario and the display of objects, an agent trained in the above way suffers from decreased decision ability. The mistake accumulates along the path, which incurs a heavy drop in successful navigation in new environments. Since strategies such as pre-exploration and beam search that exploit extra clues in a new scene are too expensive for a deployed robot, we propose a mechanism of episodic scene memory to balance accuracy and efficiency. Fig.~\ref{fig:framework} provides an overview of the proposed ESceme mechanism. By retrieving episodic memory for the $K$ views at Step $t$, ESceme replaces the vanilla encoded observations with enhanced representations for cross-modal encoding and action prediction, i.e., \textcircled{1}$\to$\textcircled{2}. In the following sections, we detail how to build the episodic scene memory and promote observations with the memory in navigation.

\subsection{Episodic scene memory construction}
We initialize the episodic memory of Scene $Y$ with an empty graph $\mathcal{G}^{(0)}_Y=(\mathcal{V}^{(0)}_Y{=}\emptyset,~ \mathcal{E}^{(0)}_Y{=}\emptyset)$ if an agent has never seen the scene. Namely, for the first instruction in Scene $Y$, an agent starts navigation with an empty episodic memory. As shown in Fig.~\ref{fig:1t1}, the start location has four neighbors and is added to $\mathcal{G}_Y$ at the end of $t{=}1$ by $\mathcal{V}^{(1)}_Y=\{V_1\}$. Node feature $\mathbf{m}_{V_1}$ is an integration of its neighbors,
\begin{align}\label{eq:pooling}
    \mathbf{m}_{V_1}=\textrm{pooling}(\mathbf{f}_{V_{1,i}}),
\end{align}
where $i{\in} \{1,2,3,4\}$ in Fig.~\ref{fig:1t1}, $\mathbf{f}_{V_{1,i}}\in \mathbb{R}^d$ is $d$-dim plain representations of the $i$-th neighbor view from the observation encoder, and $\mathbf{m}_{V_1}\in\mathbb{R}^d$. The pooling function can be either \textit{max} or \textit{mean} pooling along the number of features. It is worth noting that obtaining $\mathbf{f}_{V_{1,i}}$ does not involve extra computation since these features have been calculated in offline feature extraction. The agent selects its right neighbor to navigate, and at the end of $t{=}2$, the visited node is added to $\mathcal{G}_Y$ by $\mathcal{V}^{(2)}_Y{=}\{V_1,V_2\},~\mathcal{E}^{(2)}_Y{=}\{e_{12}\}$, with node feature $m_{V_2}$ calculated similarly as Eq.~(\ref{eq:pooling}). We set all edges $e_{jk}{=}1$. 

While following the first instruction, the agent updates its episodic scene memory $\mathcal{G}_Y$ accordingly, i.e., the green nodes and edges in Figs.~\ref{fig:1t2} and \ref{fig:1t5}. At the end of $t=5$, $\mathcal{V}_Y^{(5)}=\{V_1,V_2,...,V_5\},~\mathcal{E}_Y^{(5)}=\{e_{12},e_{23},e_{34},e_{45}\}$. When the agent is directed to the second instruction in Scene $Y$, its memory in previous visits is preserved in $\mathcal{G}_Y$ and is updated at the end of each time step accordingly as Figs.~\ref{fig:2t2} and \ref{fig:2t3} demonstrate. In Fig.~\ref{fig:2t4}, since the agent's location A has been added to ESceme in conducting the first instruction, there is no update to $\mathcal{G}_Y$. The agent stores episodic memory for each scene separately in similar ways. Therefore, we omit the subscript $Y$ for simplicity.

\subsection{ESceme navigation by candidate enhancing}\label{sec:ce}
In addition to information from instruction, current observation, and route history, an agent can refer to its episodic scene memory in decision-making at each step. 

Since the node representation in ESceme integrates information within the neighborhood, it is expected to envision the agent with a bigger picture of the current location. Therefore, we devise a candidate-enhancing (CE) mechanism to improve navigation. A flowchart of CE is shown in Fig.~\ref{fig:framework}. Faced with $K$ candidate views at Step $t$, the agent retrieves their representations $\mathbf{m}_k,~k\in \{1,...,K\}$ from episodic memory $\mathcal{G}^{(t-1)}$,
\begin{align}
    \mathbf{m}_k=\left\{ \begin{array}{ll}
        \mathbf{m}_{V_j} & \textrm{if the } k \textrm{-th view is } V_j \in \mathcal{V}^{(t-1)} \\
        \mathbf{0} & \textrm{otherwise.}
    \end{array} \right.
\end{align}
Then the Fusion block integrates the ESceme representations with the plain features $\{\mathbf{f}_1,...,\mathbf{f}_K\}$ to produce enhanced candidate viewpoints,
\begin{align}
    \mathbf{o}_k=\textrm{MLP}([\mathbf{m}_k;\mathbf{f}_k]),
\end{align}
where $[\cdot;\cdot]$ denotes concatenation along feature dimension. The MLP function is a two-layer non-linear projection from $\mathbb{R}^{2d}$ to $\mathbb{R}^d$. Following~\cite{chen2021history,zhao2022target}, type embedding that distinguishes visual and linguistic signals, navigable embedding that indicates the navigability of each candidate view, and orientation encoding are added to $\mathbf{o}_k$. A zero vector $\mathbf{o}_s\in \mathbb{R}^d$ is appended as the feature for STOP action.

Finally, together with encoded history features, the enhanced candidate representations $\{\mathbf{o}_1,...,\mathbf{o}_K,\mathbf{o}_s\}$ are input to the cross-modal encoder to merge linguistic information from encoded text features. The agent predicts the distribution of action $a_t$ via a two-layer non-linear Predictor block,
\begin{align}
    P(a_t{=}k{\in}\{1,...,K,s\}) = \frac{e^{\mathrm{MLP}(\mathbf{o}'_k\odot \mathbf{x}'_{cls})}}{\sum_{j\in\{1,...,K,s\}}e^{\mathrm{MLP}(\mathbf{o}'_j\odot \mathbf{x}'_{cls})}},
\end{align}
where $\odot$ is element-wise multiplication of two vectors $\mathbf{o}'_k$ and $\mathbf{x}'_{cls}$ $\in\mathbb{R}^d$, and the two-layer non-linear MLP block maps the result to a scalar $\in \mathbb{R}$. Following~\cite{tan2019learning,chen2021history}, we train the framework end-to-end by a mixture of Imitation Learning and Reinforcement Learning (A2C~\cite{mnih2016asynchronous}) loss,
\begin{align}
    \mathcal{L}={-}\alpha\sum_{t=1}^{T^*}\log P(a_t{=}a_t^*)-\sum_{t=1}^T\log P(\tilde{a}_t)(r_t{-}v_t),
\end{align}
where $T^*$ and $T$ are the length of the annotated route and predicted path, respectively. $\tilde{a}_t$ is sampled action. $r_t$ is the discount reward, and $v_t$ is the state value given by a two-layer (MLP) critic network.

\section{Experiments} \label{sec:exp}
\subsection{Experimental setup}

\noindent\textbf{Datasets and metrics.} We conduct experiments on the following three VLN tasks for evaluation.

\noindent(1) Short-horizon with fine-grained instructions. R2R\footnote{\url{https://github.com/peteanderson80/Matterport3DSimulator}}~\cite{anderson2018vision} constructs on Matterport3D~\cite{chang2017matterport3d} and has 7,189 direct-to-goal trajectories with an average of 10m. Each path is associated with three instructions of 29 words on average. The train, val seen, val unseen, and test unseen splits include 61, 56, 11, and 18 houses, respectively. 

\noindent(2) Long-horizon with fine-grained instructions. R4R\footnote{\url{https://github.com/google-research/google-research/tree/master/r4r}}~\cite{jain2019stay} is generated by joining existing trajectories in R2R with others that start close by where they end. Compared to R2R, it has longer paths and instructions and reduced shortest-path bias. The train, val seen, and val unseen have 233,613, 1,035, and 45,162 samples, respectively.

\noindent(3) Vision-dialog navigation. CVDN\footnote{\url{https://github.com/mmurray/cvdn/}}~\cite{jain2019stay} requires an agent to navigate given a target object and a dialog history. It has 7k trajectories and 2,050 navigation dialogs, where the paths and language contexts are also longer than those in R2R. The train, val seen, val unseen, and test splits contain 4,742, 382, 907, and 1,384 instances, respectively.

Following standard criteria~\cite{chen2021history,anderson2018vision,anderson2018evaluation}, we evaluate the R2R dataset with Trajectory Length (TL), Navigation Error (NE), Success Rate (SR), and Success weighted by Path Length (SPL). TL is the average length of an agent's navigation route in meters, NE is the mean shortest path distance between the agent's stop location and the target, and SR measures the ratio of navigation that stops less than three meters from the goal. SPL normalizes SR by the ratio between the path length of ground truth and the navigated, which balances accuracy and efficiency and becomes the key metric for the R2R dataset. We adopt three additional metrics, Coverage weighted by Length Score (CLS), normalized Dynamic Time Warping (nDTW), and Success weighted by nDTW (SDTW), to assess path fidelity on the R4R dataset. As for vision-dialog navigation on CVDN, the primary evaluation metric is Goal Progress (GP) in meters.

\noindent\textbf{Implementation details.} We adopt the encoders from~\cite{chen2021history} in comparison by default, where the text, history, and cross-modal encoders have nine, two, and four transformer layers, respectively. Features of single views are extracted offline using finetuned ViT-B/16 released by~\cite{chen2021history}. For a fair comparison, we set the feature dimension $d{=}768$, the ratio of imitation learning loss $\alpha{=}0.2$, and train the ESceme framework for 100K iterations on each dataset with a batch size of 8 and a learning rate of 1e-5. All the experiments run on a single NVIDIA V100 GPU. We adopt max pooling and single-run by default in comparison with other methods, and provide the results of mean pooling and inferring twice in ablation studies and supplementary material, with qualitative examples and failure cases included. 

For Reinforcement Learning, the action space is restricted to navigable locations (loosely equal to viewpoints) from each node, which is implemented by first predicting the log probability distribution over all the K viewpoints plus a STOP token and then setting the non-navigable ones as -inf. The policy is given by $\pi(a_t|\{o'_i\}_1^s,\{h'_i\}_1^{t-1},\{x'_i\}_{cls}^L)$, and sampling is conducted according to the restricted log probability. For Imitation Learning, the shortest path planner provides expert demonstrations, which are directly available from the simulator.

\noindent\textbf{Fair comparison.} We consider deploying an agent in new environments to execute \textit{a series of} language instructions. Admittedly, this definition is slightly different from existing methods, whereas it 1) preserves the original setup of \textit{unseen} environment, i.e., the agent never sees the environment before deployment, and 2) is more practical in real scenarios, e.g., housework robots. 
Meanwhile, the proposed episodic memory leads to initialization change: the agent conducts the first episode with empty memory and the following episodes with its own estimates. The comparisons we made in the paper aim to verify the superiority of the proposed episodic memory instead of just showing an instantiation of ESceme surpassing its counterparts. This way, we inevitably compare it with existing path memory since this is a novel memory mechanism.  
Our \textit{inter-episode} memory requires no extra time or computation while maintaining partial episodic memory via initialization, which is worth further exploration.

The directly available location ID, which we use to retrieve enhanced features for the current node, is universally adopted by 1) implicit path-level memory methods (e.g., HAMT) to retrieve accessible candidates, and 2) explicit path-level memory methods (e.g., EGP, SSM) to extend action space. Our comparisons introduce the essential signal, i.e., scan id as environment index, to be fair in showing the superiority of episodic memory over path-level memory, where \textit{unseen} scenes refer to those never appearing in training/validation. In discrete environments, the usage of location ID inevitably leads to ``the agent knowing the current location is exactly something it sees before''. The proposed ESceme can easily extend to continuous settings by combining with waypoint prediction methods (e.g., CWP~\cite{hong2022bridging}) that surpass most semantic-map-based approaches. Moreover, the proposed episodic memory mechanism can transfer to continuous scenes by maintaining a global map via the widely studied visual SLAM.

\setlength{\tabcolsep}{6.8pt}
\begin{table*}
  \centering
  \begin{tabular}{@{}l|cccc|cccc|cccc}
    \toprule
    & \multicolumn{4}{c|}{Validation Seen} & \multicolumn{4}{c|}{Validation Unseen} & \multicolumn{4}{c}{Test Unseen}  \\
    \multirow{-2}{4em}{Method}
    & TL & NE$\downarrow$ & SR$\uparrow$ & SPL$\uparrow$ & TL & NE$\downarrow$ & SR$\uparrow$ & SPL$\uparrow$ & TL & NE$\downarrow$ & SR$\uparrow$ & SPL$\uparrow$ \\ \midrule
    Seq2Seq~\cite{anderson2018vision} & 11.33 & 6.01 & 39 & - & 8.39 & 7.81 & 22 & \cellcolor[gray]{0.94}- & 8.13 & 7.85 & 20 & \cellcolor[gray]{0.94}18 \\
    SF~\cite{fried2018speaker} & - & 3.36 & 66 & - & - & 6.62 & 35 & \cellcolor[gray]{0.94}- & 14.82 & 6.62 & 35 & \cellcolor[gray]{0.94}28 \\
    AcPercep~\cite{wang2020active} & 19.7 & 3.20 & 70 & 52 & 20.6 & 4.36 & 58 & \cellcolor[gray]{0.94}40 & 21.6 & 4.33 & 60 & \cellcolor[gray]{0.94}41 \\
    PRESS~\cite{li2019robust} & 10.57 & 4.39 & 58 & 55 & 10.36 & 5.28 & 49 & \cellcolor[gray]{0.94}45 & 10.77 & 5.49 & 49 & \cellcolor[gray]{0.94}45 \\
    SSM~\cite{wang2021structured} & 14.7 & 3.10 & 71 & 62 & 20.7 & 4.32 & 62 & \cellcolor[gray]{0.94}45 & 20.4 & 4.57 & 61 & \cellcolor[gray]{0.94}46 \\
    EnvDrop~\cite{tan2019learning} & 11.00 & 3.99 & 62 & 59 & 10.70 & 5.22 & 52 & \cellcolor[gray]{0.94}48 & 11.66 & 5.23 & 51 & \cellcolor[gray]{0.94}47 \\
    OAAM~\cite{qi2020object} & 10.20 & - & 65 & 62 & 9.95 & - & 54 & \cellcolor[gray]{0.94}50 & 10.40 & - & 53 & \cellcolor[gray]{0.94}50 \\
    AuxRN~\cite{zhu2020vision} & - & 3.33 & 70 & 67 & - & 5.28 & 55 & \cellcolor[gray]{0.94}50 & - & 5.15 & 55 & \cellcolor[gray]{0.94}51 \\
    PREVALENT~\cite{hao2020towards} & 10.32 & 3.67 & 69 & 65 & 10.19 & 4.71 & 58 & \cellcolor[gray]{0.94}53 & 10.51 & 5.30 & 54 & \cellcolor[gray]{0.94}51 \\
    RelGraph~\cite{hong2020language} & 10.13 & 3.47 & 67 & 65 & 9.99 & 4.73 & 57 & \cellcolor[gray]{0.94}53 & 10.29 & 4.75 & 55 & \cellcolor[gray]{0.94}52 \\
    NvEM~\cite{an2021neighbor} & 11.09 & 3.44 & 69 & 65 & 11.83 & 4.27 & 60 & \cellcolor[gray]{0.94}55 & 12.98 & 4.37 & 58 & \cellcolor[gray]{0.94}54 \\
    NvEM+SEvol~\cite{chen2022reinforced} & 11.97 & 3.56 & 67 & 63 & 12.26 & 3.99 & 62 & \cellcolor[gray]{0.94}57 & 13.40 & 4.13 & 62 & \cellcolor[gray]{0.94}57 \\
    CSAP~\cite{wu2022cross} & 11.29 & 2.80 & 74 & 70 & 12.59 & 3.72 & 65 & \cellcolor[gray]{0.94}59 & 13.30 & 4.06 & 62 & \cellcolor[gray]{0.94}57 \\
    RecBERT~\cite{hong2021vln} & 11.13 & 2.90 & 72 & 68 & 12.01 & 3.93 & 63 & \cellcolor[gray]{0.94}57 & 12.35 & 4.09 & 63 & \cellcolor[gray]{0.94}57 \\
    ADAPT~\cite{lin2022adapt} & 11.39 & 2.70 & 74 & 69 & 12.33 & 3.66 & 66 & \cellcolor[gray]{0.94}59 & 13.16 & 4.11 & 63 & \cellcolor[gray]{0.94}57 \\
    HOP~\cite{qiao2022hop} & 11.26 & 2.72 & 75 & 70 & 12.27 & 3.80 & 64 & \cellcolor[gray]{0.94}57 & 12.68 & 3.83 & 64 & \cellcolor[gray]{0.94}59 \\
    TDSTP~\cite{zhao2022target} & - & 2.34 & 77 & 73 & - & 3.22 & 70 & \cellcolor[gray]{0.94}63 & - & 3.73 & 67 & \cellcolor[gray]{0.94}61 \\
    HAMT~\cite{chen2021history} & 11.15 & 2.51 & 76 & 72 & 11.46 & 3.62 & 66 & \cellcolor[gray]{0.94}61 & 12.27 & 3.93 & 65 & \cellcolor[gray]{0.94}60 \\
    VLN-SIG~\cite{li2023improving} & - & - & - & - & - & - & 68 & \cellcolor[gray]{0.94}62 & - & - & 65 & \cellcolor[gray]{0.94}60 \\
    KERM~\cite{li2023kerm} & 12.16 & \textbf{2.19} & \textbf{80} & \textbf{74} & 13.54 & 3.22 & 72 & \cellcolor[gray]{0.94}61 & 14.60 & 3.61 & 70 & \cellcolor[gray]{0.94}59 \\
    GridMM~\cite{wang2023gridmm} & - & - & - & - & 13.27 & \textbf{2.83} & \textbf{75} & \cellcolor[gray]{0.94}\textbf{64} & 14.43 & \textbf{3.35} & \textbf{73} & \cellcolor[gray]{0.94}62 \\
     \midrule
    ESceme (Ours) & 10.65 & 2.57 & 76 & 73 & 10.80 & 3.39 & 68 & \cellcolor[gray]{0.94}\textbf{64} & 11.89 & 3.77 & 66 & \cellcolor[gray]{0.94}\textbf{63} \\
    \bottomrule
  \end{tabular}
  \caption{Comparison with state-of-the-art methods on R2R dataset. ESceme (Ours) is built with HAMT~\cite{chen2021history} architecture by default.}
  \label{tab:r2r}
\end{table*}

\setlength{\tabcolsep}{0.5pt}
\begin{table}
  \centering
  \begin{tabular}{@{}lcccccc@{}}
    \toprule
    Method & NE\scriptsize{$\downarrow$} & SR\scriptsize{$\uparrow$} & SPL\scriptsize{$\uparrow$} & CLS\scriptsize{$\uparrow$} & nDTW\scriptsize{$\uparrow$} & SDTW\scriptsize{$\uparrow$} \\ \midrule
    SF~\cite{fried2018speaker} & 8.47 & 24 & 12 & 30 & - & - \\
    EnvDrop~\cite{tan2019learning} & - & 29 & - & 34 & - & 9 \\
    PTA~\cite{cornia2019perceive} & 8.25 & 24 & 10 & 37 & 32 & 10 \\
    RCM~\cite{wang2019reinforced} & 8.08 & 26 & 21 & 35 & 30 & 13 \\
    SSM~\cite{wang2021structured} & 8.27 & 32 & - & 53 & 39 & 19 \\
    NvEM~\cite{an2021neighbor} & 6.85 & 38 & 28 & 41 & 36 & 20 \\
    NvEM+SEvol~\cite{chen2022reinforced} & 6.90 & 39 & 29 & 41 & 36 & 20 \\
    RelGraph~\cite{hong2020language} & 7.43 & 36 & 26 & 41 & 47 & 34 \\
    EGP~\cite{deng2020evolving} & 8.00 & 30.2 & - & 44.4 & 37.4 & 17.5 \\
    TDSTP~\cite{zhao2022target} & 6.32 & 43.3 & 40.6 & 46.4 & 42.1 & 25.5 \\
    RecBERT~\cite{hong2021vln} & 6.67 & 43.6 & - & 51.4 & 45.1 & 29.9 \\
    CSAP~\cite{wu2022cross} & 6.21 & 43.0 & - & 58.6 & 51.9 & 31.5 \\
    HAMT~\cite{chen2021history} & 6.09 & 44.6 & 40.6 & 57.7 & 50.3 & 31.8 \\ \midrule
    ESceme (Ours) & \textbf{5.84} & \textbf{45.6} & \textbf{43.2} & \textbf{62.7} & \textbf{55.7} & \textbf{34.7} \\
    \bottomrule
  \end{tabular}
  \caption{Comparison on the val unseen split of R4R dataset.}
  \label{tab:r4r}
\end{table}

\setlength{\tabcolsep}{0.5pt}
\begin{table}[ht]
  \centering
  \begin{tabular}{@{}lc>{\columncolor[gray]{0.94}}c>{\columncolor[gray]{0.94}}c}
    \toprule
    \rowcolor{white}
    Method & Val Seen & Val Unseen & Test Unseen  \\ \midrule
    Seq2Seq~\cite{anderson2018vision} & 5.92 & 2.10 & 2.35 \\
    PREVALENT~\cite{hao2020towards} & - & 3.15 & 2.44 \\
    CMN~\cite{zhu2020visionb} & 7.05 & 2.97 & 2.95 \\
    VISITRON~\cite{shrivastava2022visitron} & 5.11 & 3.25 & 3.11 \\
    HOP~\cite{qiao2022hop} & - & 4.41 & 3.24 \\ 
    SCoA~\cite{zhu2021self} & 7.11 & 2.85 & 3.31 \\
    MT-RCM+EnvAg~\cite{wang2020environment} & 5.07 & 4.65 & 3.91 \\
    HAMT~\cite{chen2021history} & 6.91 & 5.13 & 5.58 \\ \midrule
    ESceme (Ours) & \textbf{8.34} & \textbf{5.42} & \textbf{5.99} \\ 
    \bottomrule
  \end{tabular}
  \caption{Resuls of Goal Process (GP) in meters on CVDN dataset.}
  \label{tab:cvdn}
\end{table}

\subsection{Comparison to state-of-the-art} \label{sec:main_exp}

\noindent\textbf{Results on R2R dataset.} Table~\ref{tab:r2r} compares the proposed ESceme with existing methods on the R2R dataset. We can see that the pretraining-finetuning paradigm (e.g., RecBERT~\cite{hong2021vln}, HAMT~\cite{chen2021history}, ADAPT~\cite{lin2022adapt}, CSAP~\cite{wu2022cross}, TDSTP~\cite{zhao2022target}) largely improves the performance of VLN in unseen environments. ESceme achieves the highest SPL on the unseen splits. It surpasses the baseline model HAMT~\cite{chen2021history} by about 5\% SPL on the validation and test unseen environments and even outperforms TDSTP~\cite{zhao2022target} that involves auxiliary training tasks. Besides, ESceme brings a relative decrease of 6.4\% and 4.1\% in NE on validation and test unseen split, respectively. The results demonstrate the efficacy of episodic scene memory in generalization to unseen scenarios with short instructions.

We also compare with the most recent works, including VLN-SIG~\cite{li2023improving}, KERM~\cite{li2023kerm}, and GridMM~\cite{wang2023gridmm}. The proxy pre-training task involved in VLN-SIG shows no advantage in unseen environments. KERM surpasses all the methods on the validation-seen split but drops much more heavily on unseen splits. GridMM achieves the highest SR and slightly lower SPL than ours in unseen scenarios yet takes a much longer trajectory length.

\setlength{\tabcolsep}{4pt}
\begin{table*}
  \centering
  \begin{tabular}{@{}lc|cccc|cccc}
    \toprule
    &  &  \multicolumn{4}{c|}{Validation Seen} & \multicolumn{4}{c}{Validation Unseen} \\
    & pooling & TL & NE$\downarrow$ & SR$\uparrow$ & SPL$\uparrow$ & TL & NE$\downarrow$ & SR$\uparrow$ & SPL$\uparrow$ \\ \midrule
    HAMT~\cite{chen2021history} & - & 11.02\scriptsize{$\pm$0.10} & 2.52\scriptsize{$\pm$0.10} & 75.0\scriptsize{$\pm$0.9} & 71.7\scriptsize{$\pm$0.7} & 11.72\scriptsize{$\pm$0.34} & 3.63\scriptsize{$\pm$0.05} & 65.7\scriptsize{$\pm$0.7} & \cellcolor[gray]{0.94}60.9\scriptsize{$\pm$0.7} \\
    ESceme & mean & 11.13\scriptsize{$\pm$0.16} & 2.59\scriptsize{$\pm$0.09} & 75.1\scriptsize{$\pm$0.7} & 71.9\scriptsize{$\pm$0.6} & 11.49\scriptsize{$\pm$0.27} & 3.50\scriptsize{$\pm$0.03} & 67.1\scriptsize{$\pm$0.5} & \cellcolor[gray]{0.94}62.3\scriptsize{$\pm$0.5} \\
    ESceme & max & 10.81\scriptsize{$\pm$0.12} & 2.60\scriptsize{$\pm$0.12} & 75.6\scriptsize{$\pm$0.4} & 72.6\scriptsize{$\pm$0.4} & 11.18\scriptsize{$\pm$0.23} & 3.44\scriptsize{$\pm$0.03} & 67.4\scriptsize{$\pm$0.5} & \cellcolor[gray]{0.94}63.2\scriptsize{$\pm$0.5} \\
    \bottomrule
  \end{tabular}
  \caption{Ablation studies of ESceme construction on R2R dataset to compare the effect of different pooling functions.}
  \label{tab:r2r_abl}
\end{table*}

\setlength{\tabcolsep}{5.8pt}
\begin{table*}
  \centering
  \begin{tabular}{@{}l|ccc|ccc|ccc}
    \toprule
    & \multicolumn{3}{c|}{Validation Seen} & \multicolumn{3}{c|}{Validation Unseen} & Params & GPU & Time \\
    \multirow{-2}{4em}{Method}
     & TL & NE$\downarrow$ & SPL$\uparrow$ & TL & NE$\downarrow$ & SPL$\uparrow$ & (MB) & (GB) & (ms) \\ \midrule
    HAMT~\cite{chen2021history} & 11.02\scriptsize{$\pm$0.10} & 2.52\scriptsize{$\pm$0.10} & 71.7\scriptsize{$\pm$0.7} & 11.72\scriptsize{$\pm$0.34} & 3.63\scriptsize{$\pm$0.05} & \cellcolor[gray]{0.94}{60.9\scriptsize{$\pm$0.7}} & 651.5 & 8.5 & 29.4 \\
    ~~~+ ESceme & 10.81\scriptsize{$\pm$0.12} & 2.60\scriptsize{$\pm$0.12} & 72.6\scriptsize{$\pm$0.4} & 11.18\scriptsize{$\pm$0.23} & 3.44\scriptsize{$\pm$0.03} & \cellcolor[gray]{0.94}{63.2\scriptsize{$\pm$0.5}} & +6.8 & +0.1 & +1.4 \\
    ~~~+ ESceme* & 10.77\scriptsize{$\pm$0.13} & 2.58\scriptsize{$\pm$0.12} & 72.8\scriptsize{$\pm$0.4} & 10.89\scriptsize{$\pm$0.14} & 3.35\scriptsize{$\pm$0.05} & \cellcolor[gray]{0.94}{64.0}\scriptsize{$\pm$0.4} & +6.8 & +0.1 & +32.2 \\ \midrule
    TDSTP~\cite{zhao2022target} & 13.09\scriptsize{$\pm$0.37} & 2.42\scriptsize{$\pm$0.08} & 70.9\scriptsize{$\pm$0.7} & 14.28\scriptsize{$\pm$0.44} & 3.22\scriptsize{$\pm$0.09} & \cellcolor[gray]{0.94}{62.1\scriptsize{$\pm$0.6}} & 657.2 & 10.5 & 46.9 \\
    ~~~+ ESceme  & 11.80\scriptsize{$\pm$0.26} & 2.34\scriptsize{$\pm$0.10} & 74.4\scriptsize{$\pm$0.6} & 13.86\scriptsize{$\pm$0.21} & 3.31\scriptsize{$\pm$0.07} & \cellcolor[gray]{0.94}{63.0\scriptsize{$\pm$0.8}} & +6.8 & +0.1 & +1.8 \\ 
    ~~~+ ESceme* & 11.83\scriptsize{$\pm$0.23} & 2.33\scriptsize{$\pm$0.08} & 74.8\scriptsize{$\pm$0.7} & 13.38\scriptsize{$\pm$0.28} & 3.21\scriptsize{$\pm$0.05} & \cellcolor[gray]{0.94}{64.3\scriptsize{$\pm$0.7}} & +6.8 & +0.1 & +50.5 \\
    \bottomrule
  \end{tabular}
  \caption{Ablation studies of navigation architectures and inferring strategies on R2R dataset. ESceme* denotes navigating with a nearly completed scene memory by first going through all the episodes. For a scene in R2R covering 92 visited nodes on average, the maximum episodic memory cost in CPU is about 1.5MB.}
  \label{tab:r2r_arc}
\end{table*}

\noindent\textbf{Results on R4R dataset.} We evaluate the proposed ESceme on the R4R dataset to examine if the generalization promotion is maintained in long-horizon navigation tasks. The results are listed in Table~\ref{tab:r4r}. Our ESceme outperforms existing state-of-the-art by a large margin, i.e., a relative improvement of 6.4\% in SPL, 7.0\% in CLS, 7.3\% in nDTW, and 9.1\% in SDTW. It indicates that ESceme improves not only navigation success but also path fidelity. Although good at carrying out short instructions, TDSTP~\cite{zhao2022target} suffers a heavy drop in long-horizon navigation regarding path fidelity compared with its baseline model HAMT~\cite{chen2021history}. It reveals that goal-related auxiliary tasks such as target prediction benefit reaching the target location but undermine the ability to follow instructions. Equipped with ESceme, an agent has a promoted ability to travel the expected route in long-horizon navigation. Besides, a consistent advantage of pretraining-based methods can be observed on this dataset.

\noindent\textbf{Results on CVND dataset.} Table~\ref{tab:cvdn} compares ESceme with state-of-the-art methods on the vision-and-dialog navigation task. CVDN provides longer instructions and trajectories than R2R and more complicated instructions than R4R. The proposed ESceme achieves the best goal process in both seen and unseen scenarios and wins first place on the leaderboard. HAMT~\cite{chen2021history} shows an obvious advantage over other pretraining-based methods such as PREVALENT~\cite{hao2020towards}, and even surpasses those counterparts specially designed for vision-and-dialog navigation, e.g., CMN~\cite{zhu2020visionb}, VISITRON~\cite{shrivastava2022visitron}, and SCoA~\cite{zhu2021self}. Our ESceme brings a relative improvement of 20.7\%, 5.7\%, and 7.3\% over the baseline HAMT~\cite{chen2021history} in val seen, val unseen, and test unseen environments, respectively.

\begin{figure*}
     \centering
     \begin{subfigure}[b]{0.33\textwidth}
         \centering
         \includegraphics[width=\textwidth]{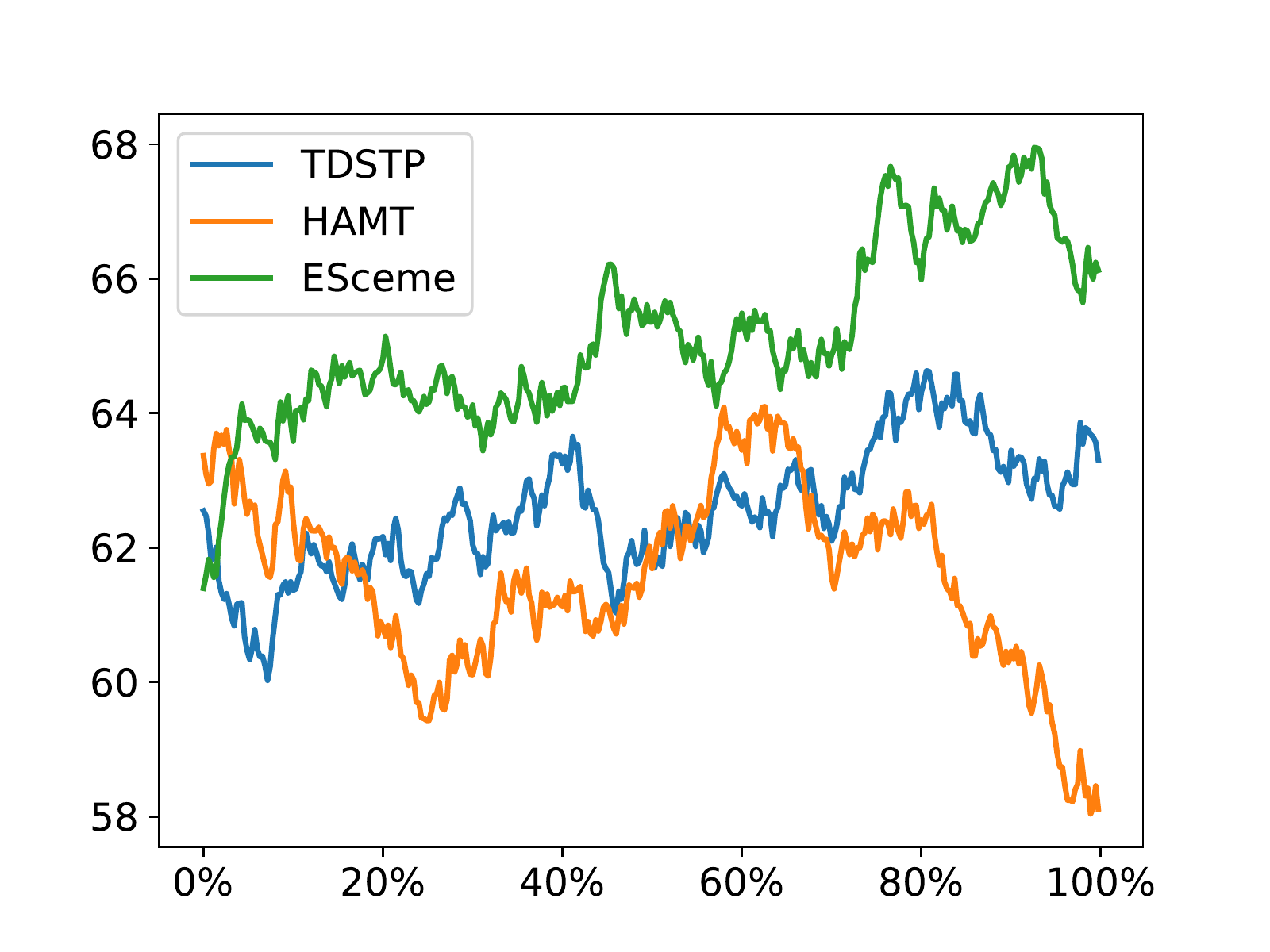}
         \caption{SPL on R2R val unseen split}
         \label{fig:r2r_spl}
     \end{subfigure}
     \hfill
     \begin{subfigure}[b]{0.33\textwidth}
         \centering
         \includegraphics[width=\textwidth]{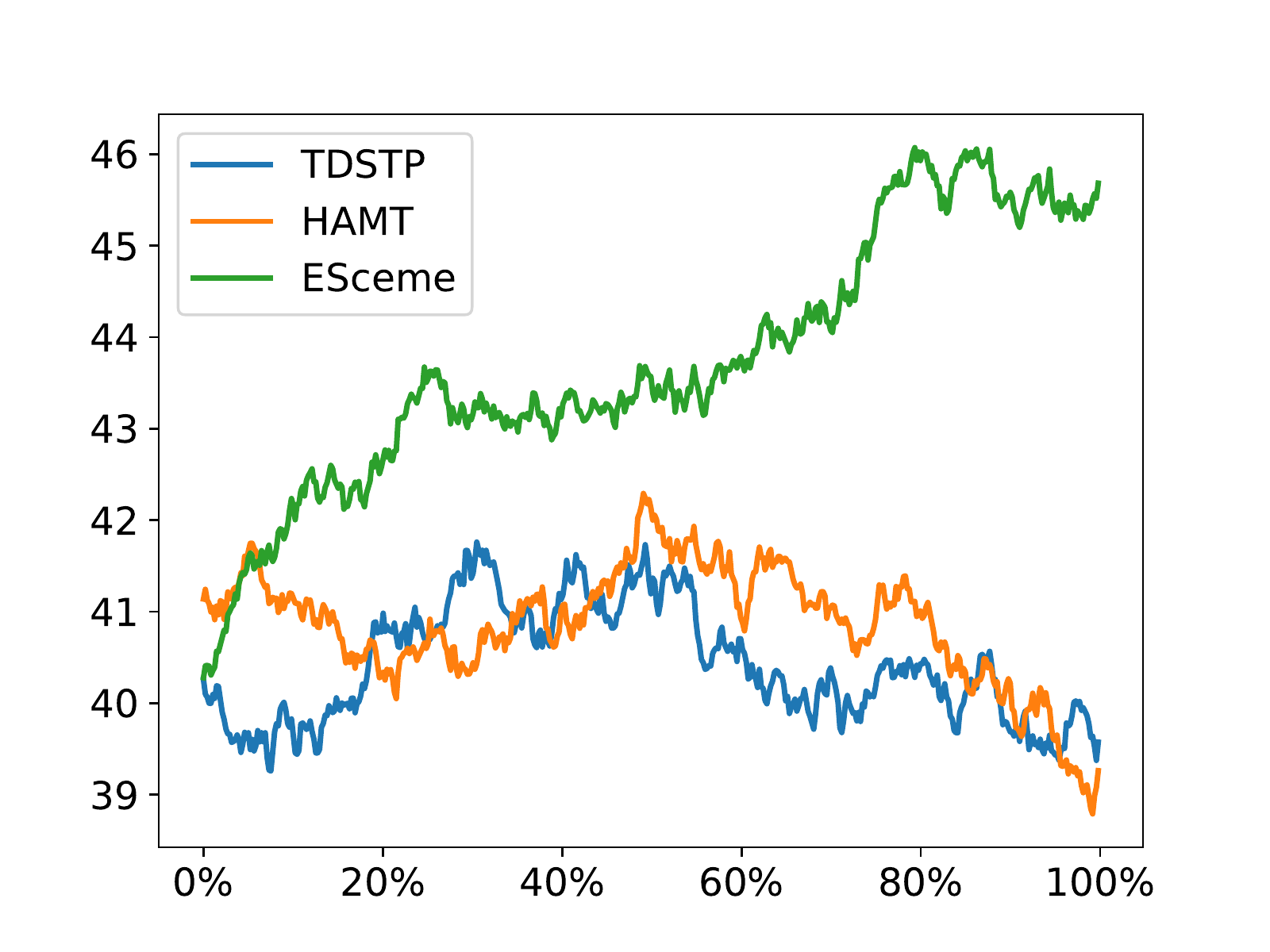}
         \caption{SPL on R4R val unseen split}
         \label{fig:r4r_spl}
     \end{subfigure}
     \hfill
     \begin{subfigure}[b]{0.33\textwidth}
         \centering
         \includegraphics[width=\textwidth]{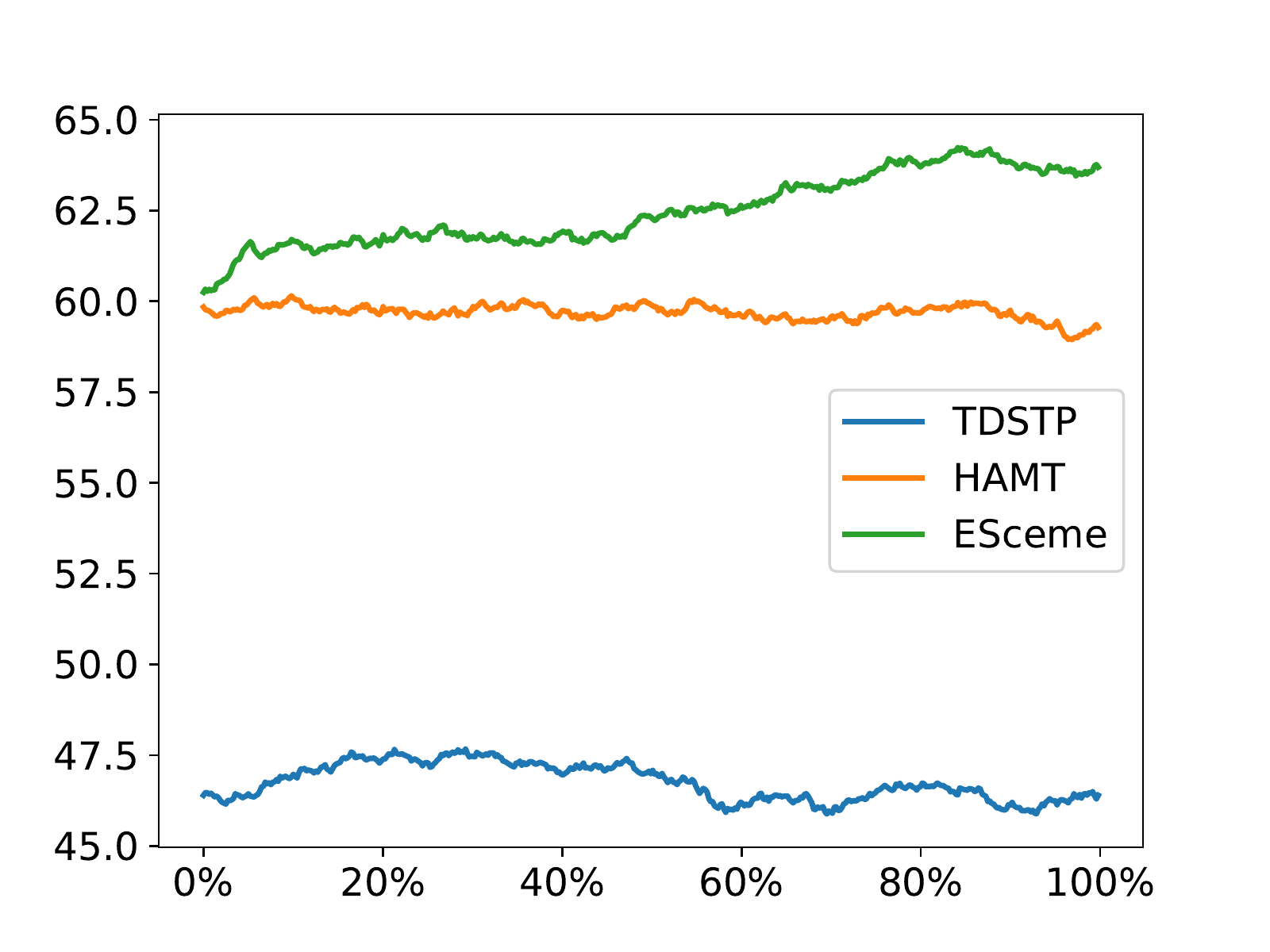}
         \caption{CLS on R4R val unseen split}
         \label{fig:r4r_cls}
     \end{subfigure}
    \caption{Navigation quality w.r.t. inferring progress. The x-axis indicates the ratio of samples tested, and the y-axis is the smoothed average of SPL or CLS. We use the default order for all the methods. Navigation with ESceme improves over time.}
    \label{fig:curves}
\end{figure*}

\subsection{Ablation studies \& analysis}
\noindent\textbf{Different ESceme constructions.} We evaluate the effect of different pooling functions in Table~\ref{tab:r2r_abl}. Candidate Enhancing with mean pooling brings a relative improvement of 2.3\% in SPL for unseen navigation and behaves similarly in seen environments. Integrated with max pooling, Candidate Enhancing further boosts the performance in unseen environments, which produces a 3.8\% relative increase compared to the HAMT~\cite{chen2021history} baseline. The results demonstrate the efficacy of the proposed Candidate Enhancing, which improves observation representations via direct injection and fusion, and max pooling, which preserves more distinguishable features of each view. Appendix A discusses a different implementation of the proposed episodic scene memory by Graph Encoding.

\noindent\textbf{Different navigation architectures \& inferring strategies.} The proposed ESceme is devised to be model-agnostic and should be compatible with any navigation network that has an observation input. To validate this property, we build ESceme upon TDSTP~\cite{zhao2022target} that achieves the highest SR on the R2R dataset and list the results in Table~\ref{tab:r2r_arc}. ESceme improves navigation in both seen and unseen environments by 4.9\% and 1.4\% in SPL, respectively. 

As introduced in Section~\ref{sec:ce}, the agent starts with an empty episodic scene memory during inference, and the memory keeps updating. If we let the agent renew its memory thoroughly by going through all the episodes and then evaluate its navigation performance, it will have a much more complete episodic memory. We present the results of ESceme* in Table~\ref{tab:r2r_arc}. We can see that the nearly completed memory further boosts the performance in unseen environments by 1.3\% and 2.1\% regarding SPL for ESceme upon HAMT~\cite{chen2021history} and TDSTP~\cite{zhao2022target}, respectively. More results of ESceme* are in supplementary material, with slighter improvements observed for longer-horizon navigation. The results demonstrate that an agent learns to assist navigation with partial and persistently updated episodic memory.

The observation that the performance of ESceme* is only slightly better than that of ESceme has two sides. On the one hand, it indicates that the agent has learned to use the dynamically accumulative episodic memory instead of working until collecting the complete memory. On the other hand, the slight gain of ESceme* indicates possible bottlenecks in the encoder/cross-encoder architecture, the frozen vision encoder, and the scale of datasets.

More effects of the proposed episodic scene memory are present in Appendix B and F. Comparison with pre-exploration methods shows that ESceme* is more robust to unseen scenarios. Ablation on graph re-initialization verifies that episodic scene memory contributes to decision-making in both seen and unseen environments. The observation in the IVLN benchmark is consistent with our discussion in Section~\ref{sec:rel_w} and our experimental results in Section~\ref{sec:main_exp}, and validates the superiority of the proposed ESceme.

\noindent\textbf{Computational efficiency.}
We present model size, GPU usage, and time cost during inference on the R2R dataset in Table~\ref{tab:r2r_arc}. Either upon HAMT~\cite{chen2021history} or TDSTP~\cite{zhao2022target}, the proposed ESceme brings about 1.0\% extra parameters and memory occupation in GPU. In a single-run setting, ESceme slightly increases the computational time by 4.8\% when built on top of HAMT. Compared with HAMT, the TDSTP baseline costs more time by 59.5\% and GPU by 23.5\%. Accordingly, our ESceme only raises the time cost by 3.8\% and almost no extra GPU consumption. With better-completed memory, ESceme* further boosts navigation performance in new environments at the expense of double the time. We can see that ESceme achieves a good trade-off between efficiency and efficacy in a single run. The proposed episodic memory mechanism consumes marginal ($\leq0.1\%$) computation and parameters. For $D$-dim features, $K$ nodes/scene, and $N$ scenes, increased cost of space and parameters are about $3.81e^{-6}{\times}DKN$ and $1.14e^{-5}{\times}D^2$, respectively.

\noindent\textbf{Order of executing instructions.} 
Since ESceme learns with dynamically updated episodic memory while conducting instructions, the order of execution has little impact on overall performance. Table~\ref{tab:order} lists navigating performance with shuffled episodes on the val unseen split in all the datasets, which indicates the stability of ESceme.
\setlength{\tabcolsep}{3pt}
\begin{table}[!h]
  \centering
  \begin{tabular}{cc|cc|c}
    \toprule
     {\scriptsize (R2R)} NE$\downarrow$ & SPL$\uparrow$ & {\scriptsize (R4R)} SPL$\uparrow$ & CLS$\uparrow$ & {\scriptsize (CVDN)} GP$\uparrow$ \\ \midrule
     3.39\scriptsize{$\pm$0.03} & 63.7\scriptsize{$\pm$0.3} & 43.2\scriptsize{$\pm$0.07} & 62.7\scriptsize{$\pm$0.1} & 5.57\scriptsize{$\pm$0.11}\\
    \bottomrule
  \end{tabular}
  \caption{$\bar{x}\pm \sigma$ scores of shuffled episodes with five random seeds on the val unseen split of the datasets.}
  \label{tab:order}
\end{table}

\noindent\textbf{Success variation during inference.} Fig.~\ref{fig:curves} compare SPL and CLS curves of different methods to visualize the variation of navigation quality in inferring progress. 
On the short-horizon navigation dataset R2R, HAMT~\cite{chen2021history} oscillates around 62 and drops in the last 1/5 progress. The decrease could result from more tough samples at the end. TDSTP~\cite{zhao2022target} presents a more stable oscillation around 62, owing to a global action space and an auxiliary goal-related task. Starting from a moderate navigation ability, an agent with ESceme benefits greatly from memory updates and maintains a high success rate with completed memory. 

On the long-horizon VLN dataset R4R, TDSTP~\cite{zhao2022target} shares a similar oscillation around 41 with HAMT~\cite{chen2021history} in SPL. TDSTP preserves a relatively more stable success rate at the cost of much lower CLS, which reveals that goal-related auxiliary task undermines the ability of instruction following. Our ESceme shows a sharp increase within the first 4/5 navigation and has remained stable since then. We attribute the excellent promotion on R4R to two reasons, 1) long-horizon navigation involves more action steps, so a slight increase in navigation ability results in a big difference in final performance; 2) the sample density of a scene from R4R is much higher than that from the R2R dataset.

\subsection{Qualitative analysis}
To intuitively demonstrate the benefit of the proposed episodic scene memory, we provide a visualization example in Fig.~\ref{fig:r2r_ex1}. It shows the panoramic views and top-down overviews of navigation. The last step of HAMT and TDSTP navigates to a visible corner of the bedroom. Instead, ESceme understands the instruction better. It takes a step to walk down to the end of the hall and then turns left to the bedroom.

A failure case of ESceme is shown in Fig.~\ref{fig:failure_r2r_1}, where the instruction is ``Leave sitting room and head towards the kitchen, turn right at living room and enter. Walk through living room to dining room and enter. Turn left and...'' After correctly predicting the first three actions, ESceme failed to enter the dining room and got lost. It indicates that the representations for the viewpoints are not distinguishable enough to capture some fine-grained difference between the dining room and the living room.

\section{Conclusion} \label{sec:conc}
In this paper, we devise the first VLN mechanism with episodic scene memory (ESceme) and propose a simple yet effective implementation via candidate enhancing. We show that an agent with ESceme improves navigation ability in short-horizon, long-horizon, and vision-and-dialog navigation. Our method outperforms the existing state-of-the-art and wins first place in the CVDN leaderboard, bringing a marginal increase in memory, parameters, and inference time. We hope this work can inspire further explorations on episodic memory in VLN and related fields, e.g., building the memory in continuous environments and with more advanced techniques such as neural SLAM.

\noindent\textbf{Limitations.} Although we show the effectiveness of the proposed episodic scene memory, there are still several limitations. First, the agent requires knowledge of environmental identity to build episodic memory for each scene. It is inevitable but supported by practical demands where an agent conducts multiple instructions in one scenario. Second, the ``location ID'' information is directly available from the simulator and the dataset, which is accurate and free of noise. For the case where location ID is unknown in advance, the episodic scene memory can be built by adding a discrete mapping process analogous to SLAM. No specific location ID is required, and the rough global position of each node can be dynamically estimated using the angle of each navigable viewpoint. Third, the architecture of a navigation agent and the training data limit the efficacy of a complete scene memory. We hope the proposed episodic scene memory can be explored in more advanced and diverse architectures.

\appendix

\section*{Appendix A: ESceme navigation by graph encoding} \label{sec:ge}
\begin{figure*}[ht]
    \centering
    \includegraphics[width=0.99\textwidth]{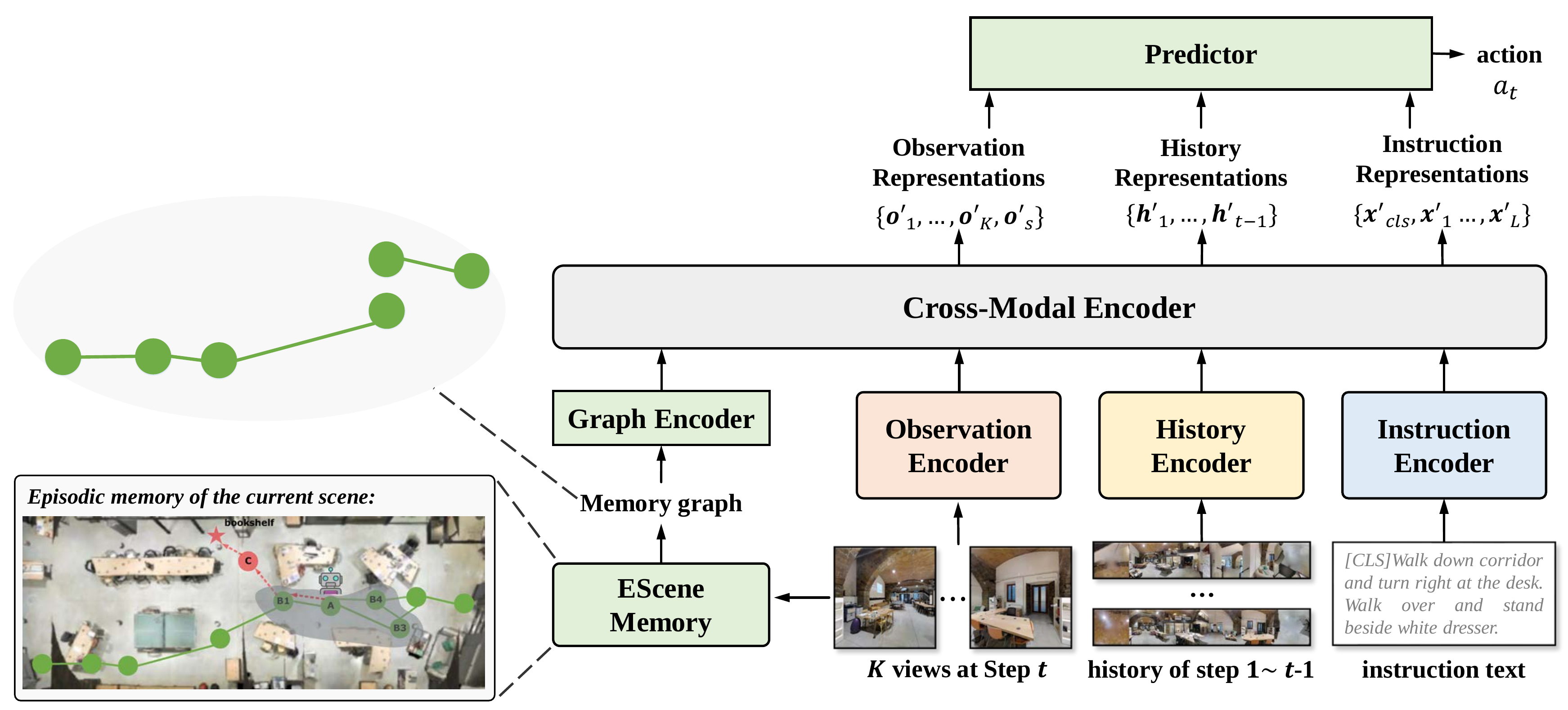}
    \caption{An overview of ESceme-assisted navigation by graph encoding. First, Episodic memory is built in the same way as that for candidate enhancing (c.f. Section 3.2). Then, the agent searches the episodic memory for the current viewpoint and obtains the memory graph by masking a local window. The encoded memory composes a separate branch to the cross-modal encoder.}
    \label{fig:ge}
\end{figure*}

Intuitively, the memory can be injected into the cross-modal encoder via a separate branch. We denote the solution as Graph Encoding (GE) and list experimental results. Fig.~\ref{fig:ge} demonstrates ESceme-assisted navigation by adding a graph encoding (GE) branch to the cross-modal encoder. At the current location where the agent stands, a local window is masked to avoid repetition with the path history from time 1 to $t{-}1$. Thus, the searched episodic memory graph includes six nodes and three edges, i.e., $\mathcal{G}^{(t-1)}=\{\mathcal{V}^{(t-1)},\mathcal{E}^{(t-1)}\}$. We adopt 3-WL GNNs~\cite{maron2019provably,dwivedi2020benchmarking} that can distinguish two non-isomorphic graphs to encode the memory graph, where the input $G\in \mathbb{R}^{n\times n\times (1+d)}$ is given by
\begin{align}
    G_{ijk}=\left\{\begin{array}{ll}
        e_{ij} & \textrm{if } k=1 \\
        m_{i} & \textrm{if } j=i \textrm{ for } k>1 \\
        0 & \textrm{otherwise,}
    \end{array} \right.
\end{align}
where $n$ is the number of nodes in $\mathcal{V}^{(t-1)}$. $m_i\in \mathbb{R}^d$ is the representation of the node $V_i$, with detailed calculations presented in Section 3.2. $e_{ij}=1$ if $V_i$ and $V_j$ are connected, else $e_{ij}=0$. The graph is encoded by
\begin{align}
    G'=[(W_1 G)\odot (W_2 G); (W_3 G)],
\end{align}
where $W_{1\sim3}{\in} \mathbb{R}^{(1+d)\times (d/2)}$ are two-layer MLPs. $\odot$ denotes element-wise multiplication and $[\cdot;\cdot]$ is the concatenation along feature dimension. The final encoded feature to the cross-modal encoder is $\sum_{i=1}^n\sum_{j=1}^n G'_{ij}{\in} \mathbb{R}^d$.

We evaluate the superiority of Candidate Enhancing over Graph Encoding and the effect of different pooling functions in Table~\ref{tab:r2r_abl_supp}. First, Graph Encoding with mean pooling slightly increases navigation success in seen environments with almost no promotion in unseen scenarios. We infer that Graph Encoding adjusts the representation of observations in cross-modal encoding and does not align well with the remaining branches to provide complementary information, resulting in a limited effect.

\setlength{\tabcolsep}{4pt}
\begin{table*}
  \centering
  \begin{tabular}{@{}lccc|cccc|cccc}
    \toprule
    &  &  &  & \multicolumn{4}{c|}{Validation Seen} & \multicolumn{4}{c}{Validation Unseen} \\
     & GE & CE & pooling & TL & NE$\downarrow$ & SR$\uparrow$ & SPL$\uparrow$ & TL & NE$\downarrow$ & SR$\uparrow$ & SPL$\uparrow$ \\ \midrule
    HAMT~\cite{chen2021history} & - & - & - & 11.02\scriptsize{$\pm$0.10} & 2.52\scriptsize{$\pm$0.10} & 75.0\scriptsize{$\pm$0.9} & 71.7\scriptsize{$\pm$0.7} & 11.72\scriptsize{$\pm$0.34} & 3.63\scriptsize{$\pm$0.05} & 65.7\scriptsize{$\pm$0.7} & \cellcolor[gray]{0.94}60.9\scriptsize{$\pm$0.7} \\
    ESceme & \cmark & \xmark & mean & 11.20\scriptsize{$\pm$0.18} & 2.56\scriptsize{$\pm$0.11} & 75.7\scriptsize{$\pm$0.9} & 72.3\scriptsize{$\pm$0.6} & 11.64\scriptsize{$\pm$0.05} & 3.60\scriptsize{$\pm$0.06} & 65.9\scriptsize{$\pm$0.5} & \cellcolor[gray]{0.94}60.9\scriptsize{$\pm$0.6}  \\
    ESceme & \xmark & \cmark & mean & 11.13\scriptsize{$\pm$0.16} & 2.59\scriptsize{$\pm$0.09} & 75.1\scriptsize{$\pm$0.7} & 71.9\scriptsize{$\pm$0.6} & 11.49\scriptsize{$\pm$0.27} & 3.50\scriptsize{$\pm$0.03} & 67.1\scriptsize{$\pm$0.5} & \cellcolor[gray]{0.94}62.3\scriptsize{$\pm$0.5} \\
    ESceme & \xmark & \cmark & max & 10.81\scriptsize{$\pm$0.12} & 2.60\scriptsize{$\pm$0.12} & 75.6\scriptsize{$\pm$0.4} & 72.6\scriptsize{$\pm$0.4} & 11.18\scriptsize{$\pm$0.23} & 3.44\scriptsize{$\pm$0.03} & 67.4\scriptsize{$\pm$0.5} & \cellcolor[gray]{0.94}63.2\scriptsize{$\pm$0.5} \\
    \bottomrule
  \end{tabular}
  \caption{Ablation studies of ESceme construction on R2R dataset. We compare the effect of graph encoding (GE) and candidate enhancing (CE), and different pooling functions.}
  \label{tab:r2r_abl_supp}
\end{table*}

\section*{Appendix B: Effects of the episodic scene memory} \label{sec:1vs2} 
We thoroughly compare navigating with progressively completed and nearly complete episodic memory on three datasets in Tables~\ref{tab:r2r_sm} and \ref{tab:r4r_sm}. ESceme conducts instructions in a single-run setting, where the agent dynamically updates memory in inference. ESceme* first goes through all the episodes to build a nearly complete memory at the beginning of the evaluation. ESceme* improves navigating in new environments by 1.6\% (SPL) on test unseen split of the R2R dataset. As for vision-dialog navigation CVDN, the improvement in val unseen and test unseen is 5.5\% and 3.0\%, respectively. On the long-horizon navigation dataset R4R, the relative increase is about 0.5\%.

Overall, ESceme* further promotes generalization to novel scenarios, indicating that ESceme benefits from the nearly complete scene memory. On the other hand, the small gap between ESceme and ESceme* shows that the agent has learned to utilize progressively completed memory in navigation.

Besides, Table~\ref{tab:pre-exp} lists the comparison with pre-exploration methods. The pre-exploration methods achieve very competitive results on val seen split while suffering from a heavier drop in unseen environments. In Table~\ref{tab:reinit}, we test ESceme with the memory graph re-initialized at every episode. The results on the R2R dataset verify that the ESceme agent indeed benefits from episodic memory for decision-making in both seen and unseen environments.

\setlength{\tabcolsep}{6pt}
\begin{table*}
  \centering
  \begin{tabular}{@{}l|cccc|cccc|cccc}
    \toprule
    & \multicolumn{4}{c|}{Validation Seen} & \multicolumn{4}{c|}{Validation Unseen} & \multicolumn{4}{c}{Test Unseen}  \\
    Method & TL & NE$\downarrow$ & SR$\uparrow$ & SPL$\uparrow$ & TL & NE$\downarrow$ & SR$\uparrow$ & SPL$\uparrow$ & TL & NE$\downarrow$ & SR$\uparrow$ & SPL$\uparrow$ \\ \midrule
    ESceme & 10.65 & 2.57 & 76 & 73 & 10.80 & 3.39 & 68 & \cellcolor[gray]{0.94}64 & 11.89 & 3.77 & 66 & \cellcolor[gray]{0.94}63 \\
    ESceme* & 10.62 & 2.57 & 76 & 73 & 10.65 & 3.36 & 68 & \cellcolor[gray]{0.94}64 & 11.77 & 3.69 & 68 & \cellcolor[gray]{0.94}64 \\ \bottomrule
    \end{tabular}
    \caption{Results of different inferring strategies on R2R dataset.}
    \label{tab:r2r_sm}
\end{table*}

\setlength{\tabcolsep}{6pt}
\begin{table*}[!h]
  \centering
  \begin{tabular}{@{}l|cccccc|ccc}
    \toprule
     & \multicolumn{6}{c|}{R4R Val Unseen} & \multicolumn{3}{c}{CVDN} \\
    Method & NE$\downarrow$ & SR$\uparrow$ & SPL$\uparrow$ & CLS$\uparrow$ & nDTW$\uparrow$ & SDTW$\uparrow$ & Val Seen & Val Unseen & Test Unseen \\
    \midrule
    ESceme & 5.84 & 45.6 & 43.2 & 62.7 & 55.7 & 34.7 & 8.34 & \cellcolor[gray]{0.94}5.42 & \cellcolor[gray]{0.94}5.99\\
    ESceme* & 5.83 & 45.8 & 43.4 & 62.9 & 55.8 & 34.8 & 8.39 & \cellcolor[gray]{0.94}5.72 & \cellcolor[gray]{0.94}6.17 \\
    \bottomrule
    \end{tabular}
    \caption{Results of different inferring strategies on R4R and CVDN datasets.}
    \label{tab:r4r_sm}
\end{table*}

\setlength{\tabcolsep}{6pt}
\begin{table*}[!ht]
  \centering
  \begin{tabular}{@{}l|cccc|cccc}
    \toprule
    & \multicolumn{4}{c|}{Validation Seen} & \multicolumn{4}{c}{Validation Unseen}  \\
    \multirow{-2}{4em}{Method}
    & TL & NE$\downarrow$ & SR$\uparrow$ & SPL$\uparrow$ & TL & NE$\downarrow$ & SR$\uparrow$ & SPL$\uparrow$ \\ \midrule
    Follower~\cite{fried2018speaker} & 10.40 & 3.68 & 65 & 62 & 9.57 & 5.20 & 52 & \cellcolor[gray]{0.94}49 \\
    Speaker~\cite{fried2018speaker} & 11.19 & 3.80 & 61 & 56 & 10.71 & 4.25 & 55 & \cellcolor[gray]{0.94}49 \\
    VLN-BERT~\cite{majumdar2020improving} & 10.28 & 3.73 & 70 & 66 & 9.60 & 4.10 & 59 & \cellcolor[gray]{0.94}55 \\
    Airbert~\cite{guhur2021airbert} & 11.09 & 2.68 & 75 & 70 & 11.78 & 4.01 & 62 & \cellcolor[gray]{0.94}56 \\
    ESceme* & 10.62 & \textbf{2.57} & \textbf{76} & \textbf{73} & 10.65 & \textbf{3.36} & \textbf{68} & \cellcolor[gray]{0.94}\textbf{64} \\
    \bottomrule
  \end{tabular}
  \caption{Comparison between ESceme* and pre-exploration methods on R2R dataset.}
  \label{tab:pre-exp}
\end{table*}

\setlength{\tabcolsep}{3pt}
\begin{table}[!h]
  \centering
  \begin{tabular}{@{}l|cccc|cccc}
    \toprule
    & \multicolumn{4}{c|}{Validation Seen} & \multicolumn{4}{c}{Validation Unseen}  \\
    \multirow{-2}{4em}{Method}
    & TL & NE$\downarrow$ & SR$\uparrow$ & SPL$\uparrow$ & TL & NE$\downarrow$ & SR$\uparrow$ & SPL$\uparrow$ \\ \midrule
    ESceme & 10.65 & 2.57 & 76 & \textbf{73} & 10.80 & 3.39 & 68 & \cellcolor[gray]{0.94}\textbf{64} \\
    re-initialize & 13.88 & 3.53 & 64 & 60 & 14.88 & 4.33 & 55 & \cellcolor[gray]{0.94}51 \\
    \bottomrule
  \end{tabular}
  \caption{Ablation of memory re-initialization on R2R dataset.}
  \label{tab:reinit}
\end{table}

\section*{Appendix C: Experiments on RxR dataset}
Results on RxR\footnote{\url{https://github.com/google-research-datasets/RxR}}~\cite{rxr} in Table~\ref{tab:rxr} indicate that longer instructions and trajectories add sufficient knowledge to the proposed episodic memory to overcome coreference challenge and promote navigation. 
\setlength{\tabcolsep}{1.7pt}
\begin{table}[ht]
  \centering
  \begin{tabular}{@{}l|cccc|cccc}
    \toprule
    & \multicolumn{4}{c|}{Validation Seen} & \multicolumn{4}{c}{Validation Unseen}  \\
    \multirow{-2}{3.4em}{Method}
    & \scriptsize{SR$\uparrow$} & \scriptsize{SPL$\uparrow$} & \scriptsize{nDTW$\uparrow$} & \scriptsize{SDTW$\uparrow$} & \scriptsize{SR$\uparrow$} & \scriptsize{SPL$\uparrow$} & \scriptsize{nDTW$\uparrow$} & \scriptsize{SDTW$\uparrow$} \\ \midrule
    HAMT & 59.4 & 58.9 & 65.3 & 50.9 & 56.5 & 56.0 & 63.1 & 48.3 \\
    ESceme & \textbf{64.8} & \textbf{61.4} & \textbf{68.8} & \textbf{56.2} & \textbf{62.0} & \textbf{58.4} & \textbf{67.6} & \textbf{53.5} \\
    \bottomrule
  \end{tabular}
  \caption{Comparison on RxR dataset. ESceme (Ours) is built with HAMT architecture by default.}
  \label{tab:rxr}
\end{table}

\setlength{\tabcolsep}{3pt}
\begin{table}[!h]
  \centering
  \begin{tabular}{@{}l|cccc|cccc}
    \toprule
    & \multicolumn{4}{c|}{Validation Seen} & \multicolumn{4}{c}{Validation Unseen}  \\
    \multirow{-2}{4em}{Method}
    & TL & NE$\downarrow$ & SR$\uparrow$ & SPL$\uparrow$ & TL & NE$\downarrow$ & SR$\uparrow$ & SPL$\uparrow$ \\ \midrule
    HAMT & 10.1 & 4.2 & 63 & 61 & 9.4 & 4.7 & 56 & \cellcolor[gray]{0.94}54 \\
    IVLN & 9.4 & 5.8 & 45 & 43 & 10.0 & 6.2 & 39 & \cellcolor[gray]{0.94}36 \\
    ESceme & 10.0 & 4.4 & 63 & 61 & 9.2 & 4.6 & 58 & \cellcolor[gray]{0.94}\textbf{56} \\
    \bottomrule
  \end{tabular}
  \caption{Comparison with IVLN~\cite{krantz2023iterative}.}
  \label{tab:ivln}
\end{table}

\section*{Appendix D: Pseudo-code implementation} \label{sec:algo}
We provide the pseudo-code of ESceme construction and candidate enhancing in Algorithm~\ref{alg:esceme}. ESceme requires easy implementation and can be integrated with any navigation networks that encode the observation.

\begin{algorithm*}[!ht]
  \algsetup{linenosize=\tiny}
  \scriptsize
\SetAlFnt{\tiny}
\SetAlCapFnt{\small}
\SetAlCapNameFnt{\small}
\SetAlgoLined
\PyCode{def update($G_Y$, ob):} \\
\Indp
\PyComment{$G_Y$: episodic memory of Scene Y} \\
\PyComment{ob: observation structure that includes information of the current viewpoint and $K$ views of a panorama} \\
\PyCode{if ob.viewpoint not in $G_Y$:} \\
\Indp
\PyCode{$G_Y$.update(ob.viewpoint), feat=pooling(ob.navigable\_feats))} \PyComment{Update nodes by pooling ViT features} \\
\PyCode{for node in ob.navigable\_views:} \\
\Indp
\PyCode{if node in $G_Y$:} \\
\Indp
\PyCode{$G_Y$.add\_edge(ob.viewpoint, node)} \PyComment{Update edges} \\
\Indm \Indm \Indm
\PyCode{return $G_Y$} \\ 
\Indm
\hfill \\
\PyCode{def candidate\_enhancing($G_Y$, ob):} \\
\Indp
\PyCode{$\mathbf{m}$ = $G_Y$.search(ob.navigable\_views).feat} \PyComment{Search memory representations for currently navigable views} \\
\PyCode{$\mathbf{o}$ = MLP(torch.cat([ob.navigable\_feats, $\mathbf{m}$], dim=1]}) \PyComment{Enhance candidate representations} \\
\PyCode{$\mathbf{o}$ = $\mathbf{o}$ + embeddings} \PyComment{Optionally add embeddings from orientation and navigable type} \\ 
\PyCode{return $\mathbf{o}$} \\ 
\Indm 
\caption{ESceme construction and candidate enhancing}
\label{alg:esceme}
\end{algorithm*}

\section*{Appendix E: Qualitative examples and failure cases} \label{sec:qualitative} 
\begin{figure*}
     \centering
     \begin{subfigure}[b]{0.98\textwidth}
         \centering
         \includegraphics[width=\textwidth]{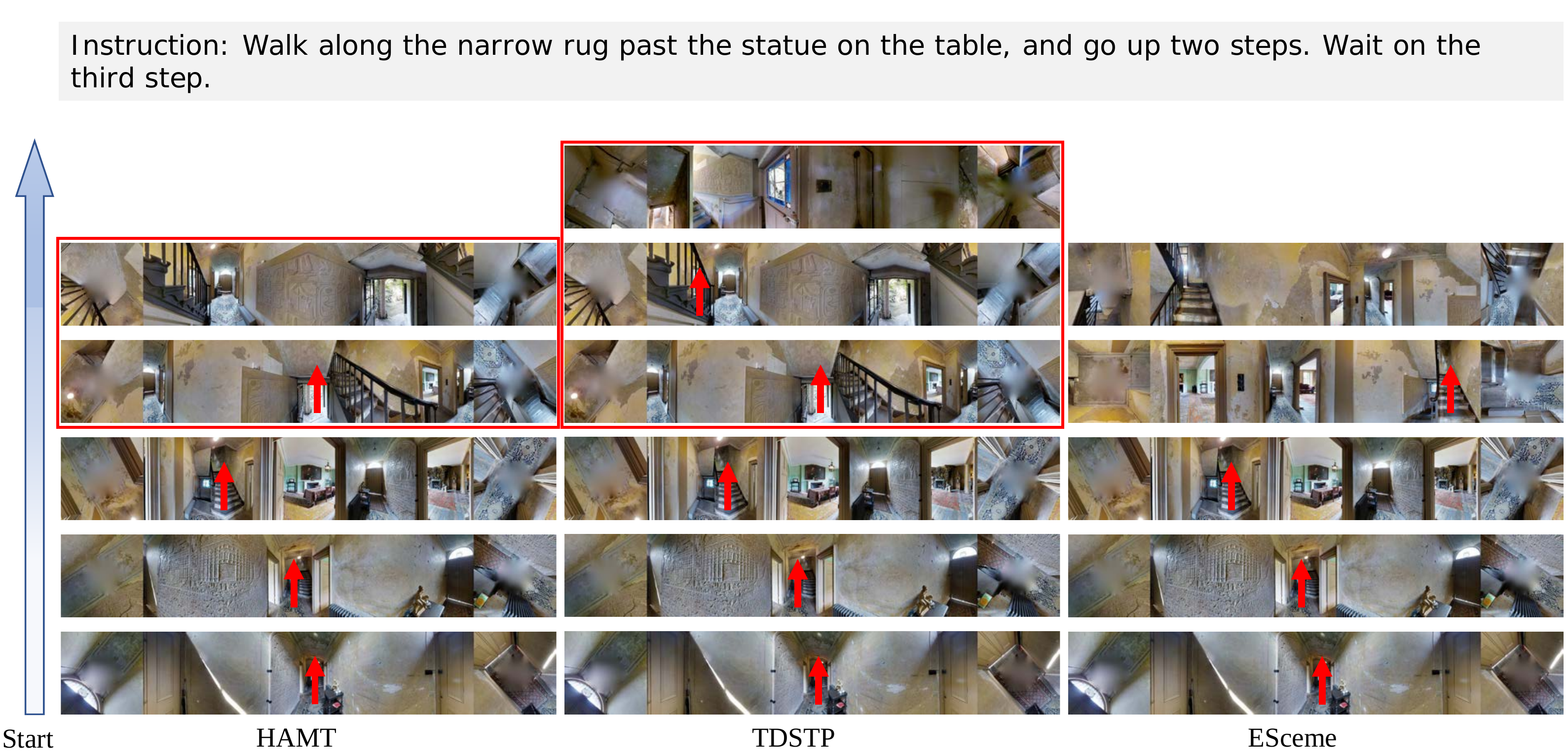}
         \label{fig:r2r_ex2_pano}
     \end{subfigure}
     \hfill
     \begin{subfigure}[b]{0.98\textwidth}
         \centering
         \includegraphics[width=\textwidth]{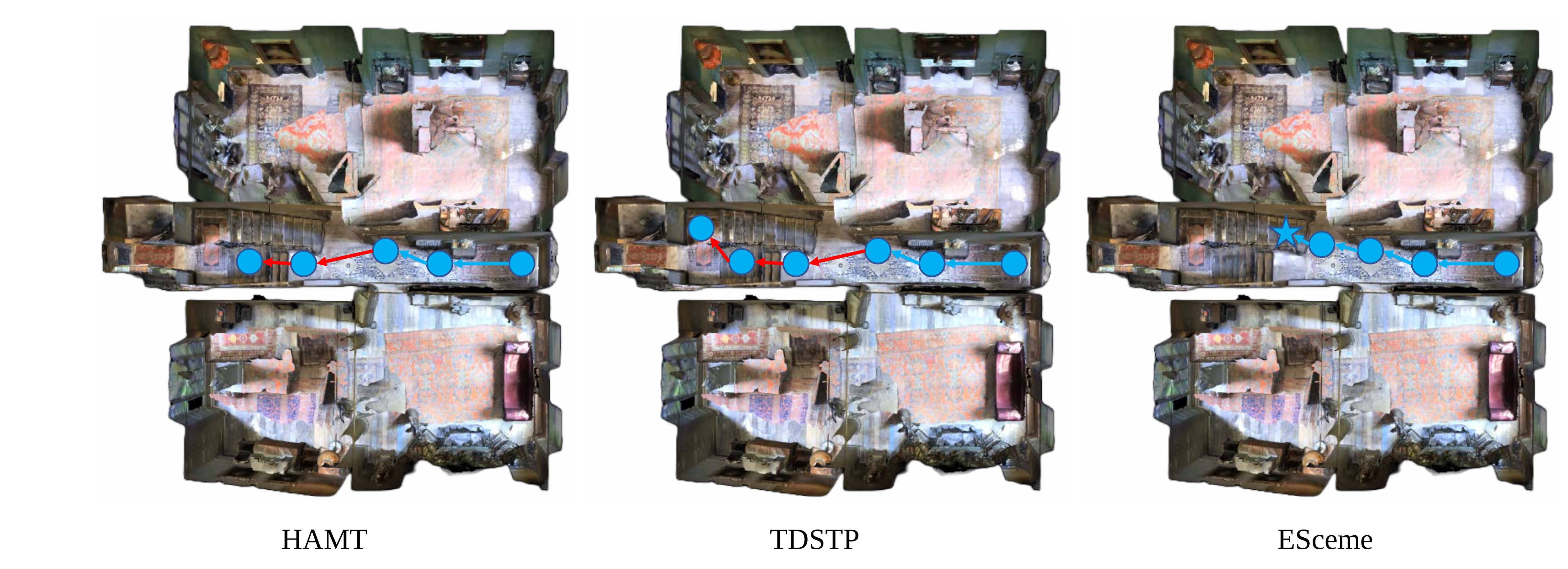}
         \label{fig:r2r_ex2_route}
     \end{subfigure}
    \caption{Panoramic views and top-down overviews of navigation. Mistakes during navigation are marked with \textcolor{red}{red} boxes for panorama and \textcolor{red}{red} arrows for top-down trajectories. The star indicates the target location. Our ESceme strictly follows the instruction ``go up two steps'' and waits on the third step.}
    \label{fig:r2r_ex2}
\end{figure*}

\begin{figure*}
     \centering
     \begin{subfigure}[b]{0.98\textwidth}
         \centering
         \includegraphics[width=\textwidth]{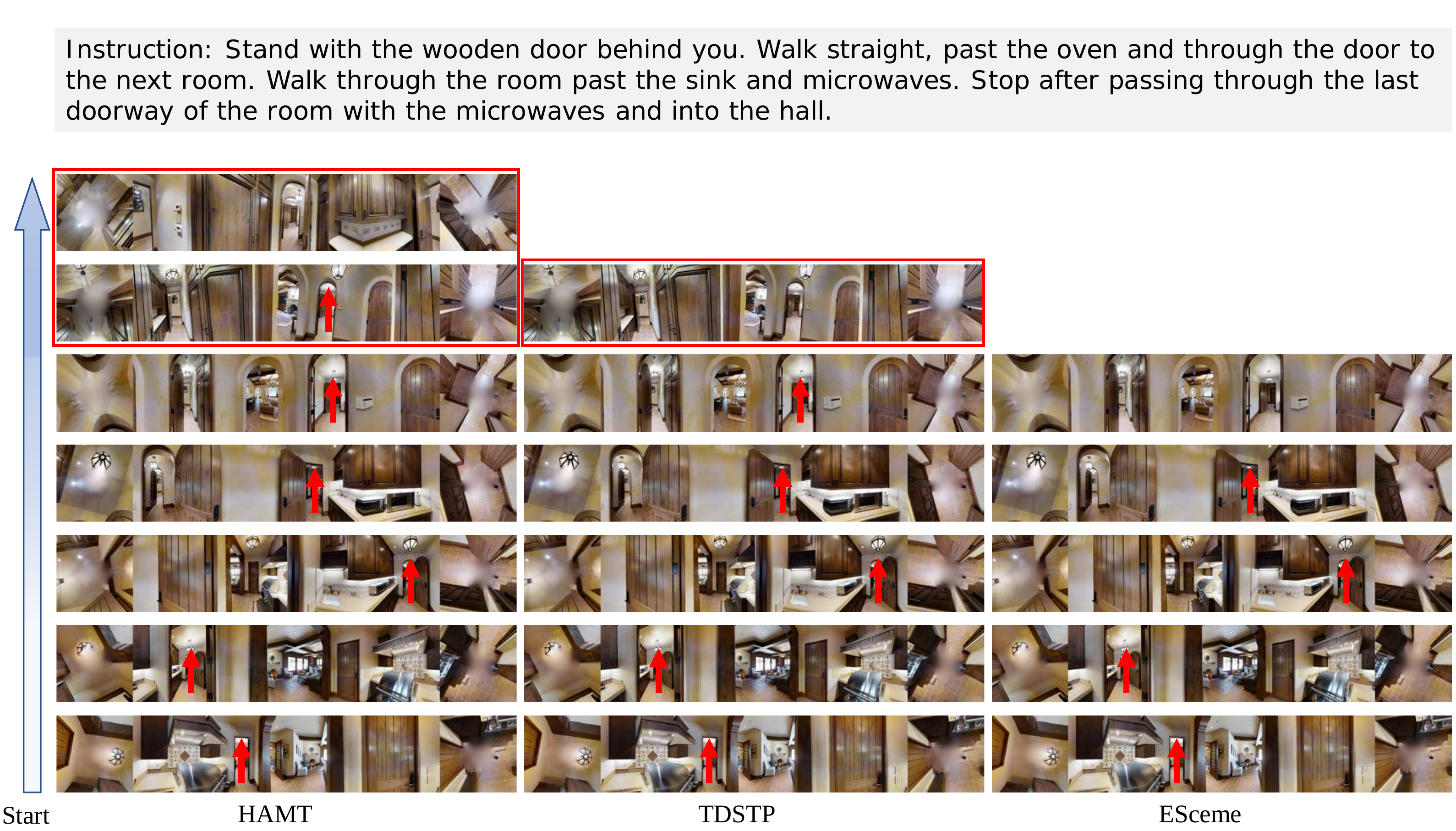}
         \label{fig:r2r_ex3_pano}
     \end{subfigure}
     \hfill
     \begin{subfigure}[b]{0.98\textwidth}
         \centering
         \includegraphics[width=\textwidth]{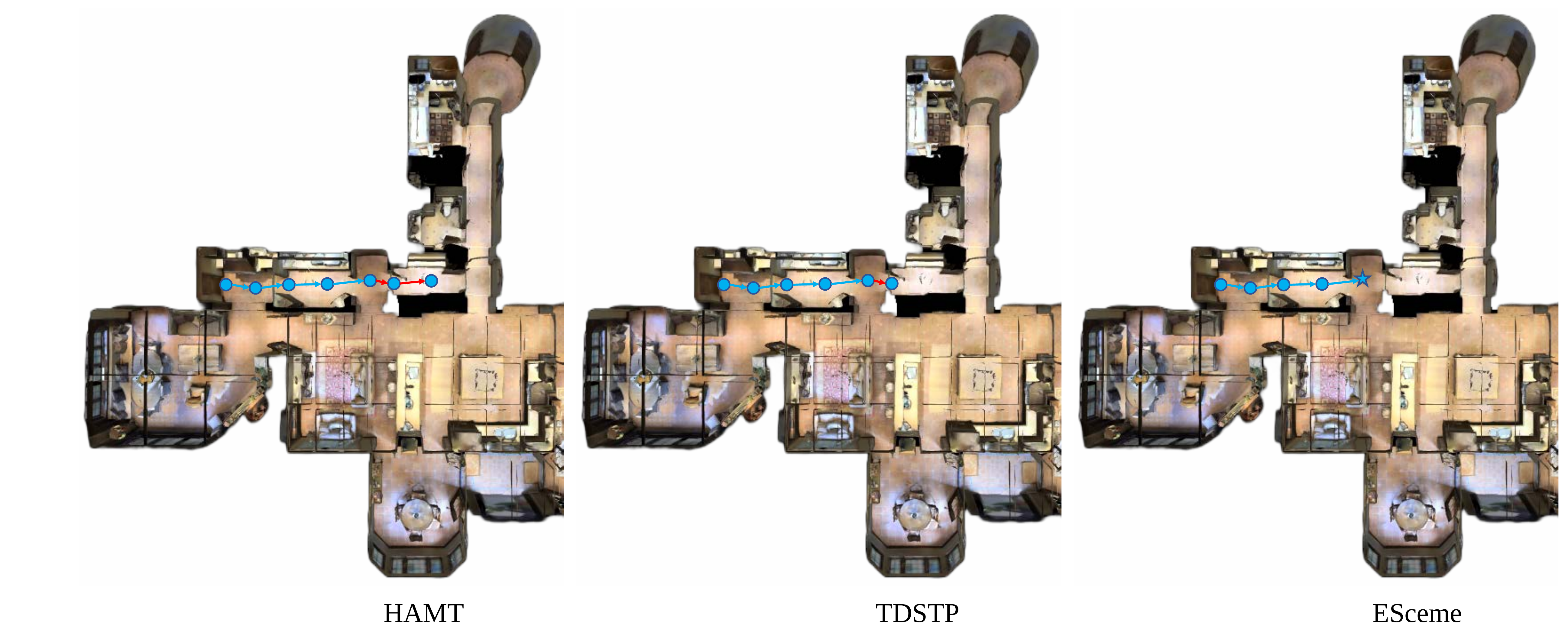}
         \label{fig:r2r_ex3_route}
     \end{subfigure}
    \caption{Panoramic views and top-down overviews of navigation. Mistakes during navigation are marked with \textcolor{red}{red} boxes for panorama and \textcolor{red}{red} arrows for top-down trajectories. The star indicates the target location. Our ESceme stops at the right place.}
    \label{fig:r2r_ex3}
\end{figure*}

\begin{figure*}
     \centering
     \begin{subfigure}[b]{0.96\textwidth}
         \centering
         \includegraphics[width=\textwidth]{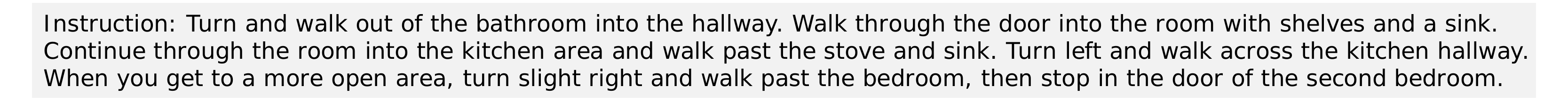}
     \end{subfigure}
     \hfill
     \vspace{0.2cm}
     \begin{subfigure}[b]{0.6\textwidth}
         \centering
         \includegraphics[width=\textwidth]{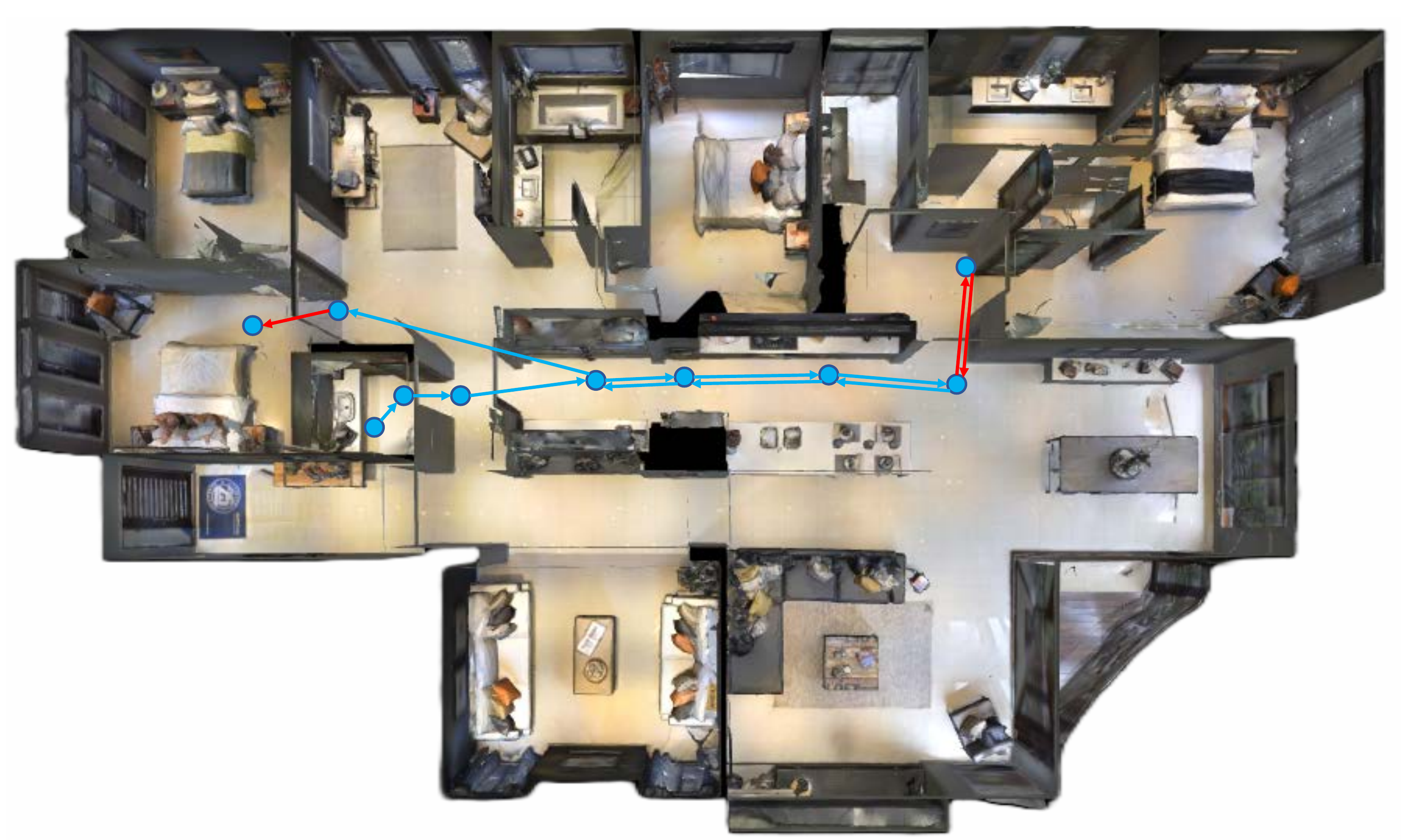}
         \caption{HAMT}
         \label{fig:r4r_ex1_hamt}
     \end{subfigure}
     \hfill
     \begin{subfigure}[b]{0.6\textwidth}
         \centering
         \includegraphics[width=\textwidth]{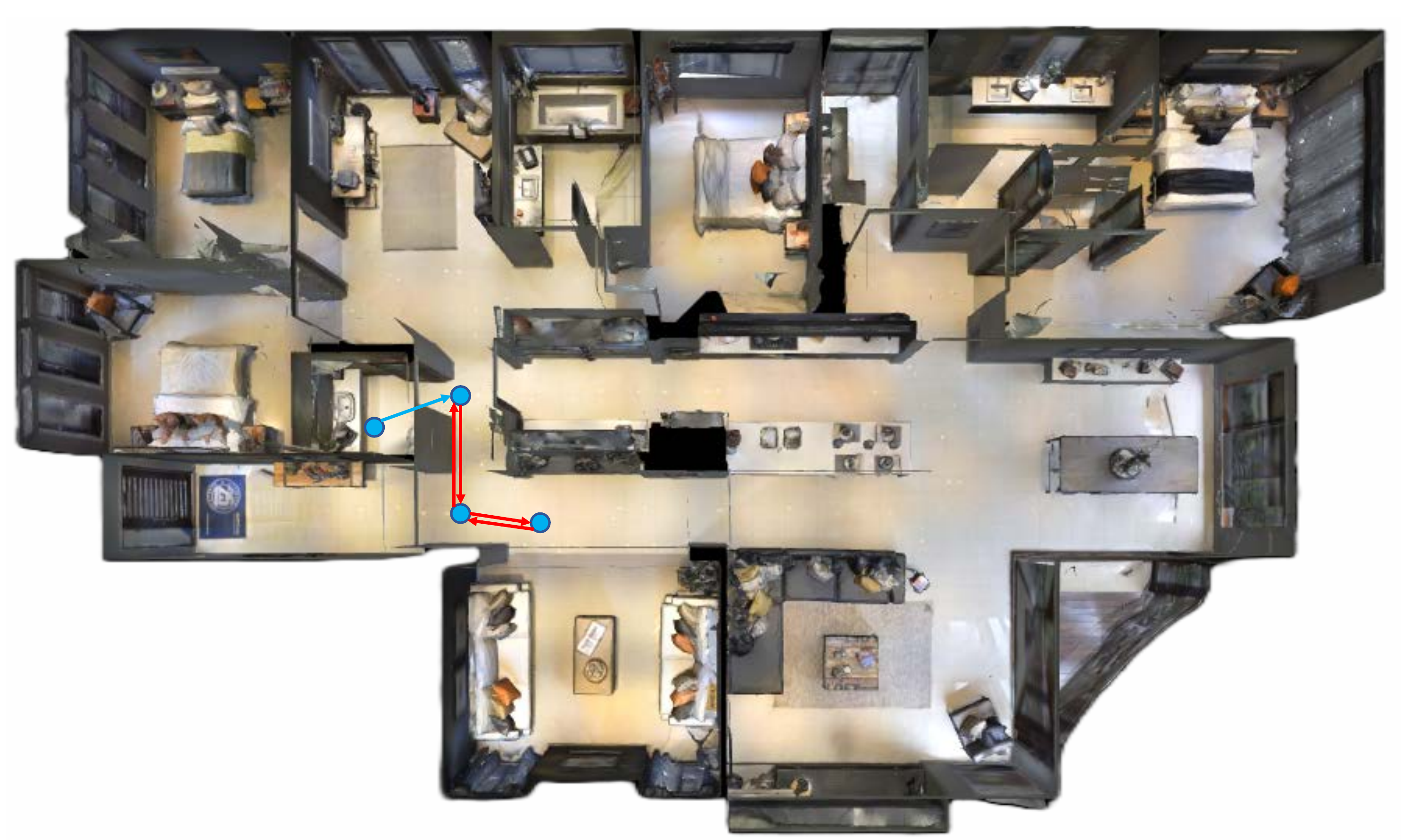}
         \caption{TDSTP}
         \label{fig:r4r_ex1_tdstp}
     \end{subfigure}
     \hfill
     \begin{subfigure}[b]{0.6\textwidth}
         \centering
         \includegraphics[width=\textwidth]{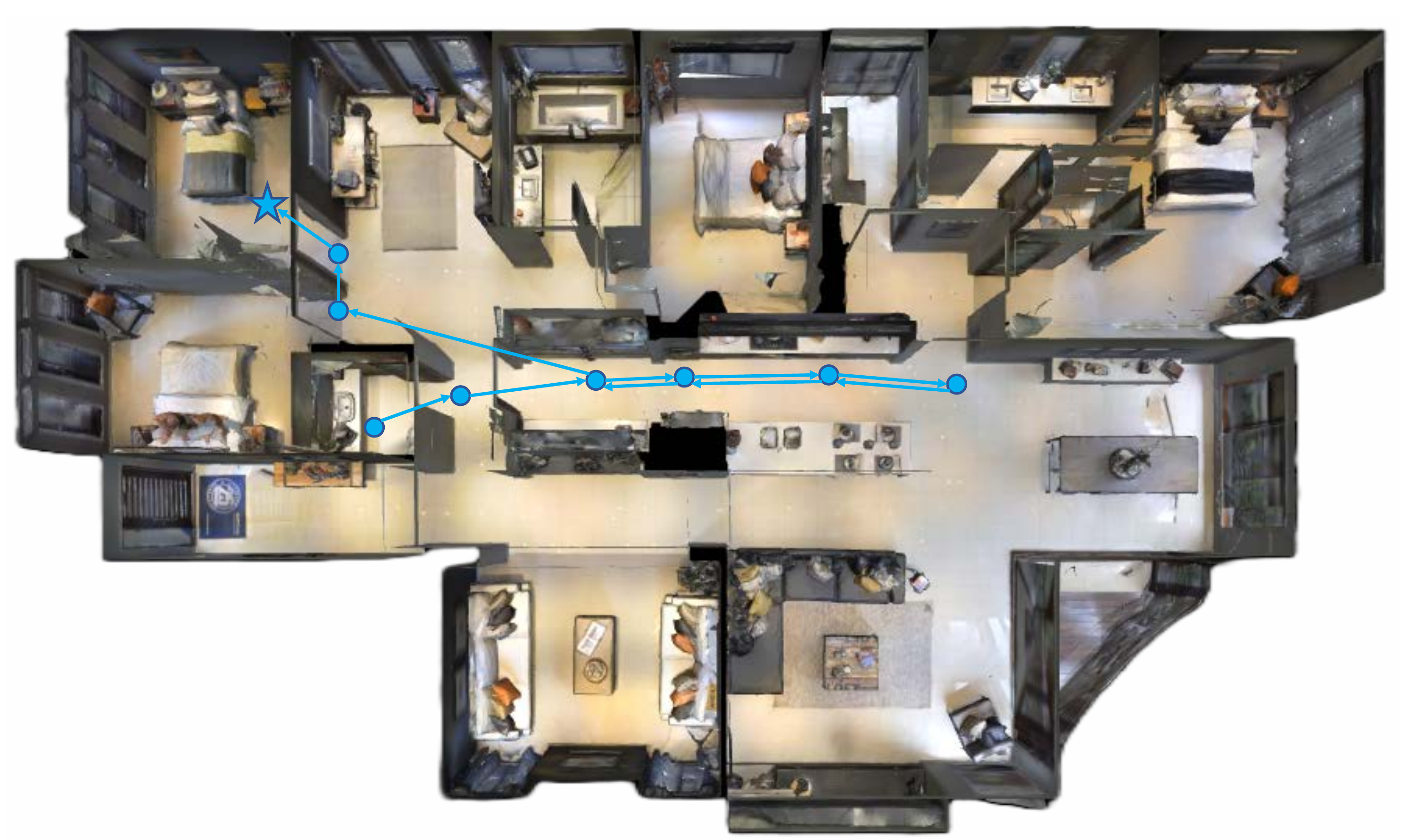}
         \caption{ESceme}
         \label{fig:r4r_ex1_esceme}
     \end{subfigure}
    \caption{The top-down trajectory of navigation. Mistakes during navigation are marked with \textcolor{red}{red}. The star indicates the target location. Our ESceme moves forward to the right place and then back and arrives at the second bathroom.}
    \label{fig:r4r_ex1}
\end{figure*}

\begin{figure*}
     \centering
     \begin{subfigure}[b]{0.96\textwidth}
         \centering
         \includegraphics[width=\textwidth]{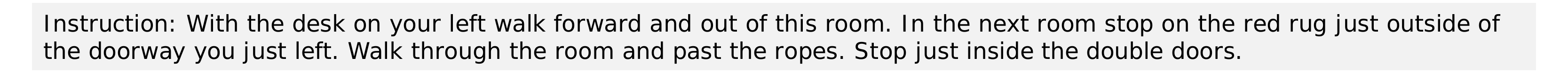}
     \end{subfigure}
     \hfill
     \vspace{0.2cm}
     \begin{subfigure}[b]{0.56\textwidth}
         \centering
         \includegraphics[width=\textwidth]{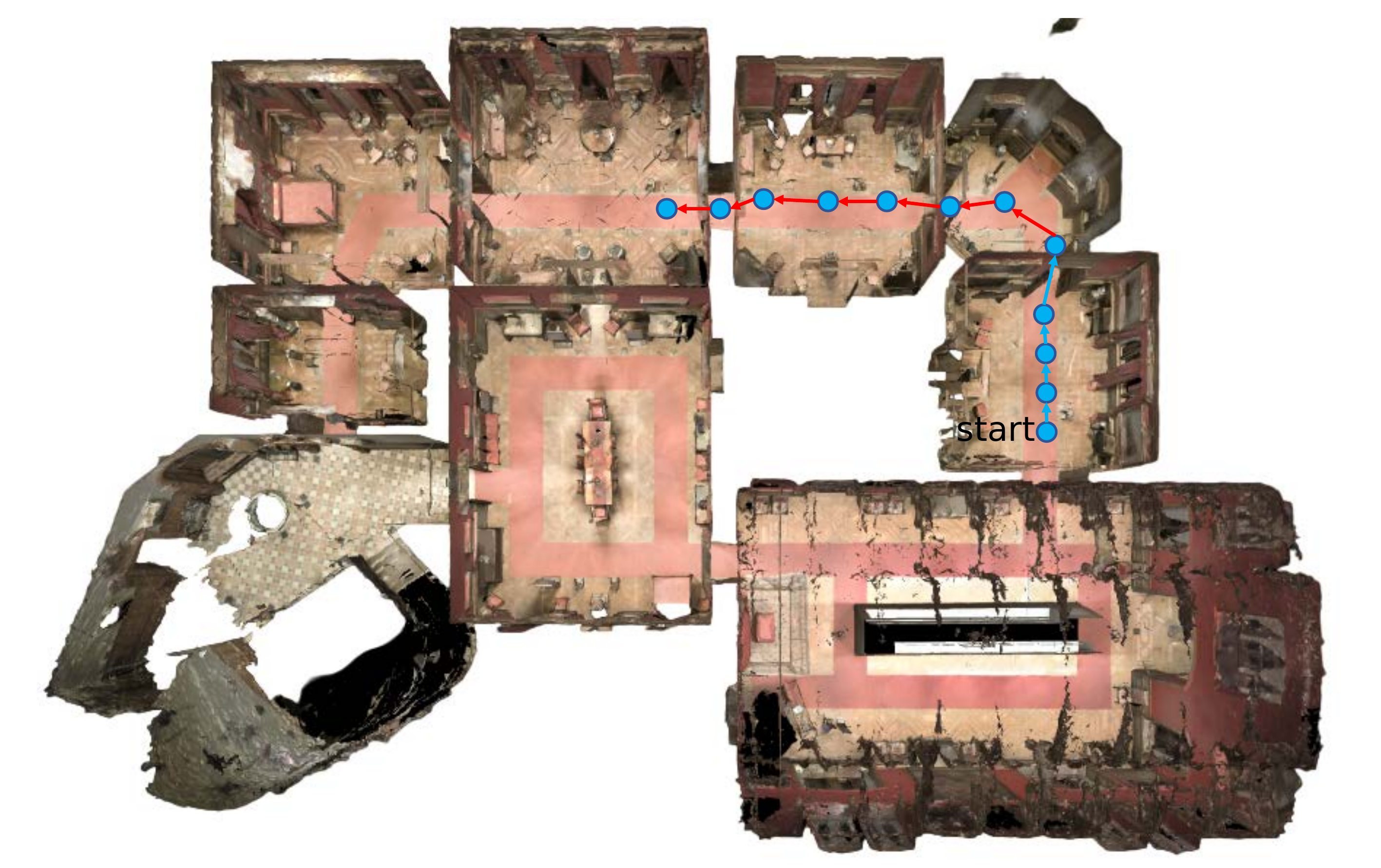}
         \caption{HAMT}
         \label{fig:r4r_ex2_hamt}
     \end{subfigure}
     \hfill
     \begin{subfigure}[b]{0.56\textwidth}
         \centering
         \includegraphics[width=\textwidth]{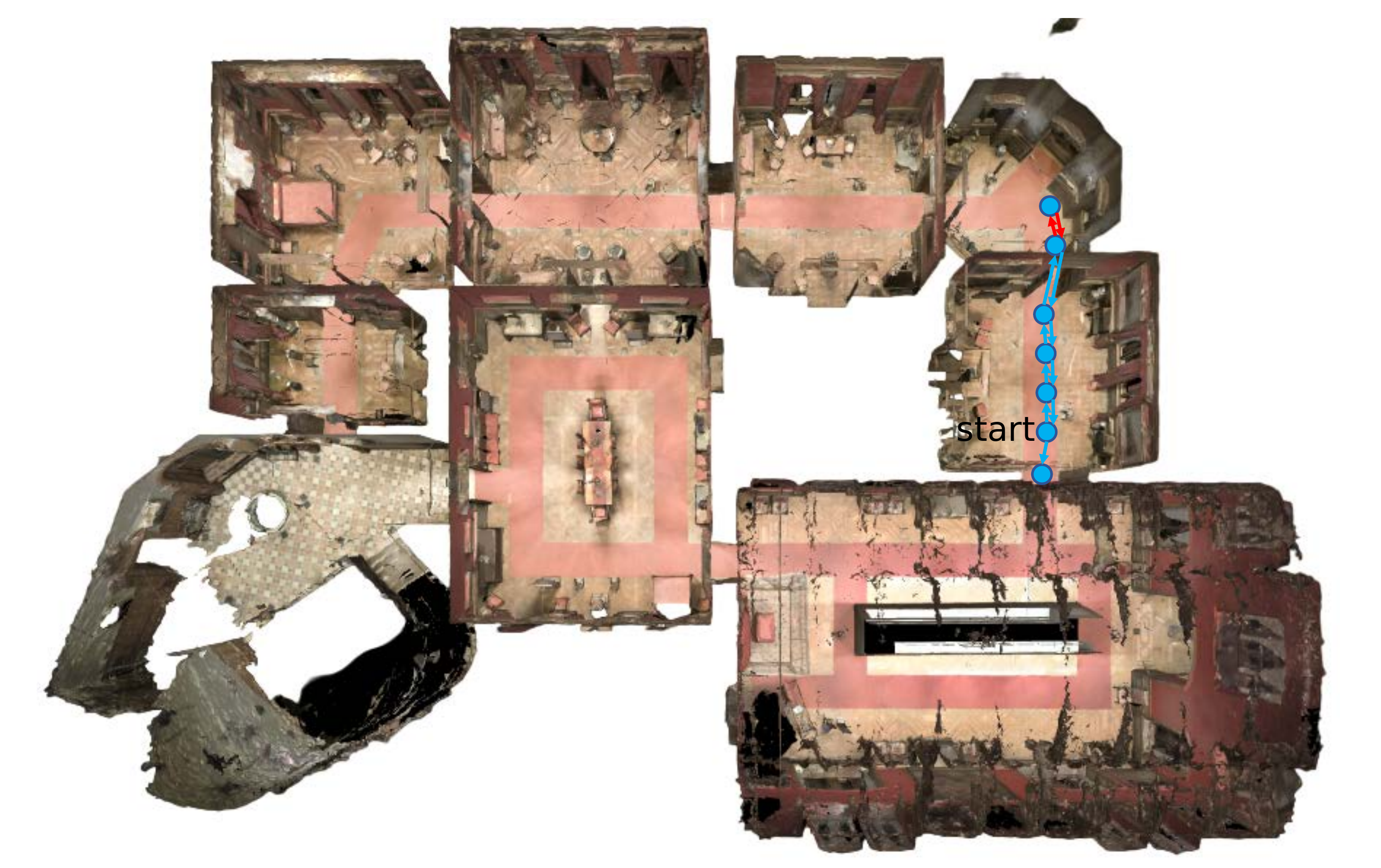}
         \caption{TDSTP}
         \label{fig:r4r_ex2_tdstp}
     \end{subfigure}
     \hfill
     \begin{subfigure}[b]{0.56\textwidth}
         \centering
         \includegraphics[width=\textwidth]{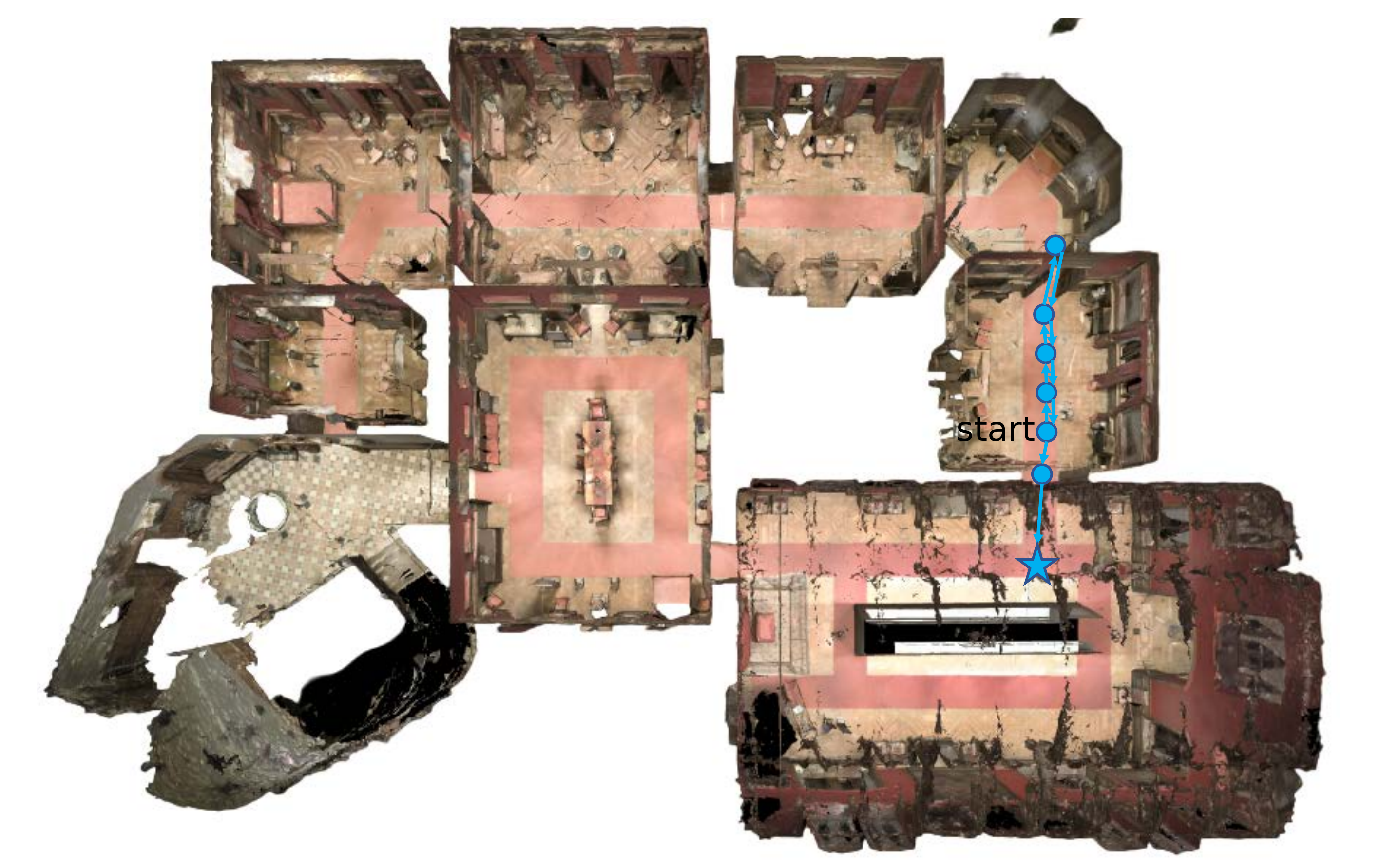}
         \caption{ESceme}
         \label{fig:r4r_ex2_esceme}
     \end{subfigure}
    \caption{The top-down trajectory of navigation. Mistakes during navigation are marked with \textcolor{red}{red}. The star indicates the target location. Our ESceme moves forward to the right place and then back and stops inside the double doors.}
    \label{fig:r4r_ex2}
\end{figure*}

We present the navigating process to provide a more intuitive comparison with HAMT~\cite{chen2021history} and TDSTP~\cite{zhao2022target}. Figs.~\ref{fig:r2r_ex2} and \ref{fig:r2r_ex3} are two navigation examples on R2R dataset, and Figs.~\ref{fig:r4r_ex1} and \ref{fig:r4r_ex2} illustrate two examples on R4R dataset. All the examples are tested in unseen environments. For short-horizon navigation, our ESceme outperforms its counterparts regarding stopping precision. For long-horizon navigation, our ESceme shows an improved ability to follow instructions that require a forward and back trip and arrives at the target location. We attribute these advantages to the episodic memory of the scenes.

\begin{figure*}
    \centering
     \begin{subfigure}[b]{0.49\textwidth}
         \centering
         \includegraphics[width=\textwidth]{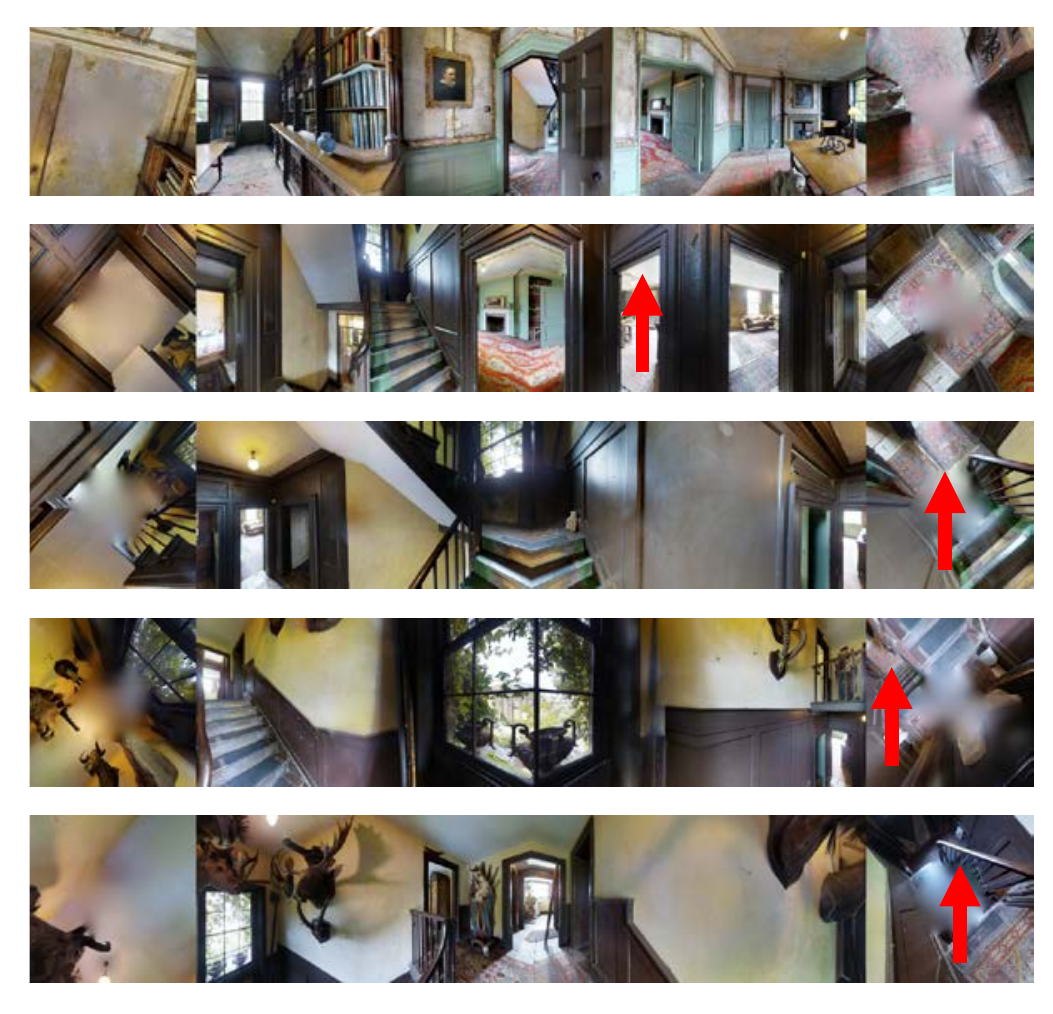}
         \caption{ground-truth trajectory}
     \end{subfigure}
     \hfill
     \vspace{0.2cm}
     \begin{subfigure}[b]{0.49\textwidth}
         \centering
         \includegraphics[width=\textwidth]{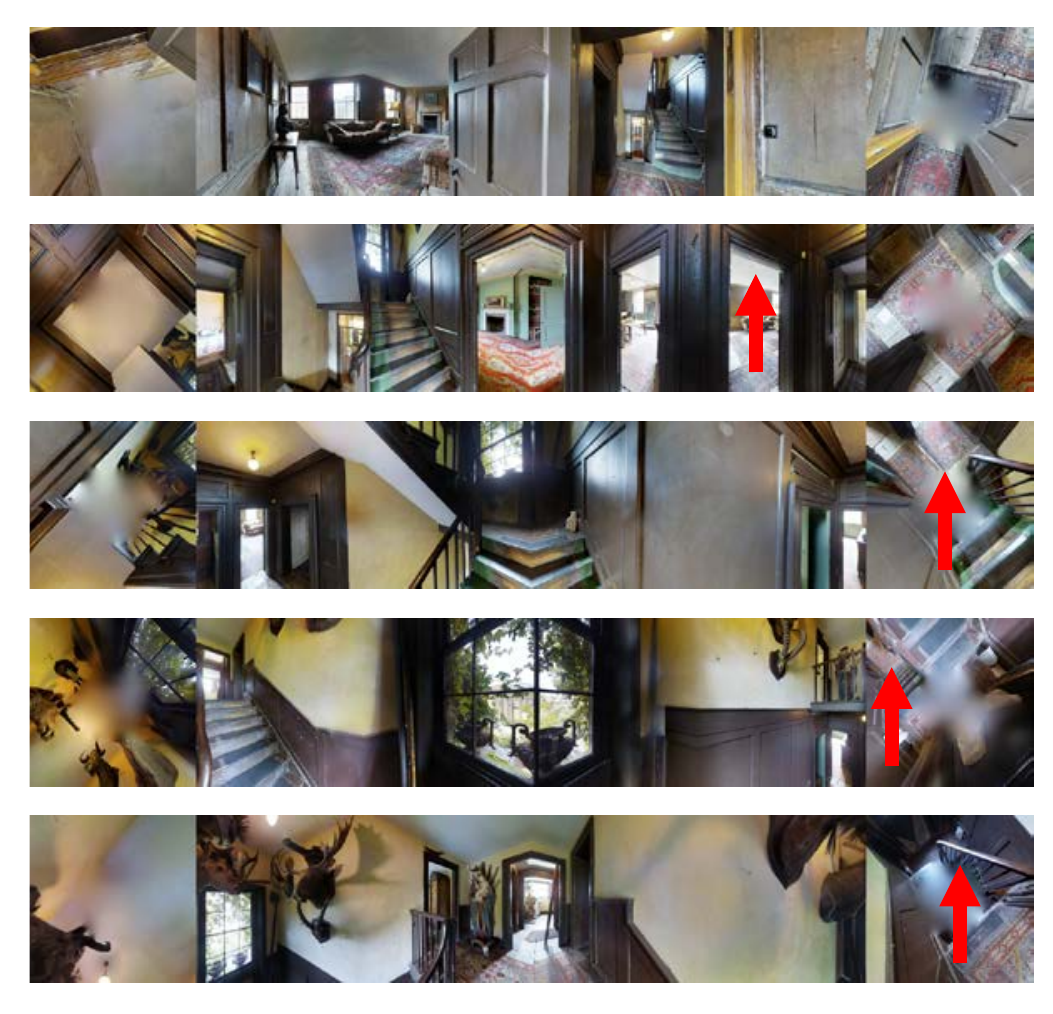}
         \caption{trajectory predicted by ESceme}
     \end{subfigure}
     \caption{Failure case in R2R val unseen split. The instruction is ``Go down the stairs. Go into the room straight ahead on the slight left. Wait there.'' ESceme succeeded in going downstairs but failed to determine the \textit{slight left} direction and entered the wrong room.}
     \label{fig:failure_r2r_2}
\end{figure*}

Figs.~\ref{fig:failure_r2r_2} showcase one more situation where ESceme failed to follow the instructions. 
The instruction is ``Go down the stairs. Go into the room straight ahead on the slight left. Wait there.'' ESceme succeeded in going downstairs but failed to determine the \textit{slight left} direction and entered the wrong room. The result indicates difficulties in understanding finer-grained instructions and distinguishing finer-grained visual observations in physical scenarios, as discussed in Section~\ref{sec:conc} Limitations.

\section*{Appendix F: Comparison with IVLN} \label{sec:ivln} 
Our setting is identical to IVLN, which reorganizes episodes into tours. Table~\ref{tab:ivln} is the direct comparison using the IVLN benchmark. IVLN decreases the performance in both seen and unseen environments, yet our ESceme promotes navigation in unseen scenarios while maintaining the performance in seen ones. The conclusion is consistent with our observations in Section~\ref{sec:exp}.

\begin{figure*}
     \centering
     \begin{subfigure}[b]{0.98\textwidth}
         \centering
         \includegraphics[width=\textwidth]{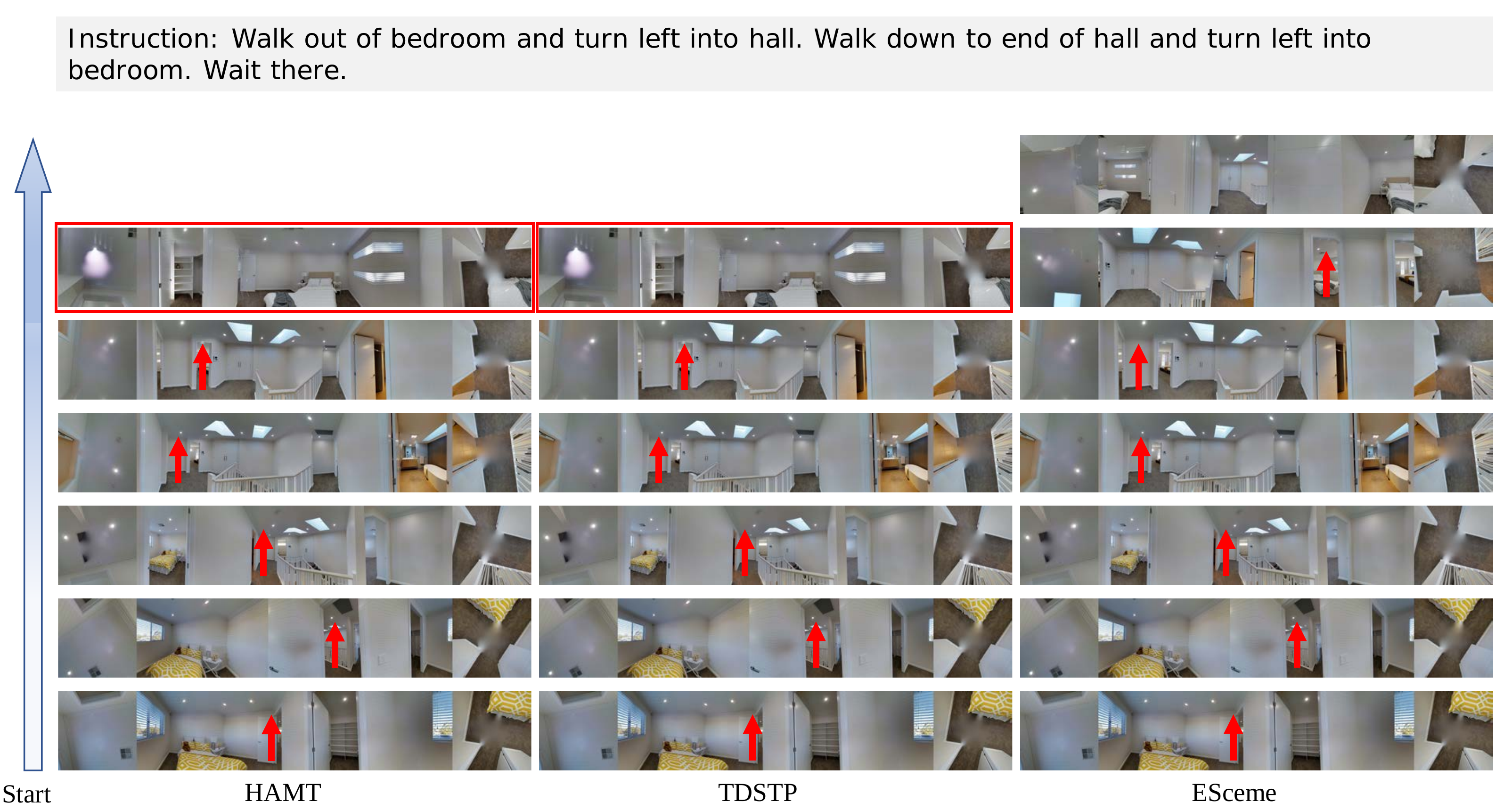}
         \label{fig:r2r_ex1_pano}
     \end{subfigure}
     \hfill
     \begin{subfigure}[b]{0.98\textwidth}
         \centering
         \includegraphics[width=\textwidth]{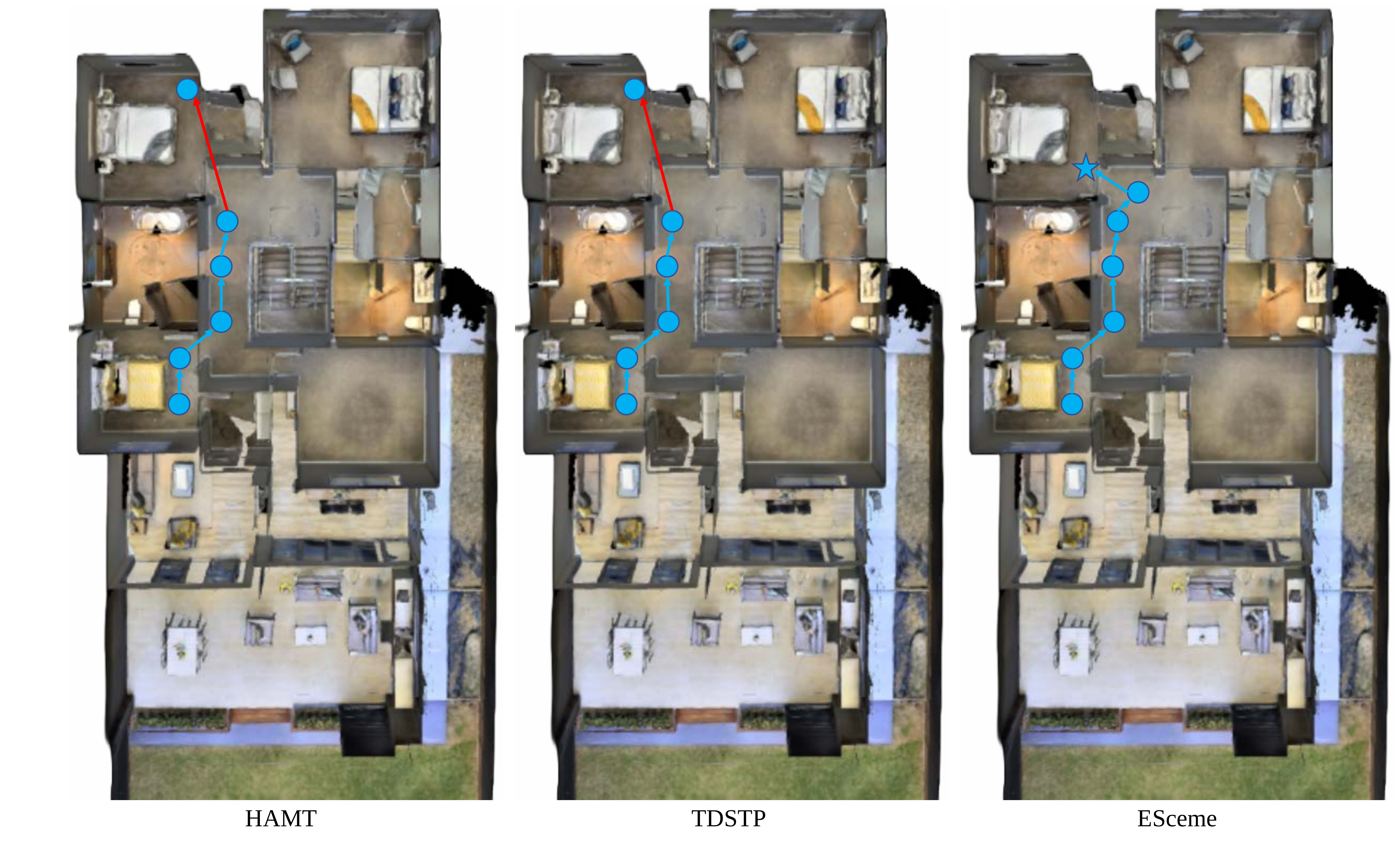}
         \label{fig:r2r_ex1_route}
     \end{subfigure}
    \caption{Panoramic views and top-down overviews of navigation. Mistakes during navigation are marked with \textcolor{red}{red} boxes for panorama and \textcolor{red}{red} arrows for top-down trajectories. The star indicates the target location. Our ESceme strictly follows the instruction ``walk down to the end of hall'' and waits at the door of the bedroom.}
    \label{fig:r2r_ex1}
\end{figure*}

\begin{figure*}
    \centering
     \begin{subfigure}[b]{0.49\textwidth}
         \centering
         \includegraphics[width=\textwidth]{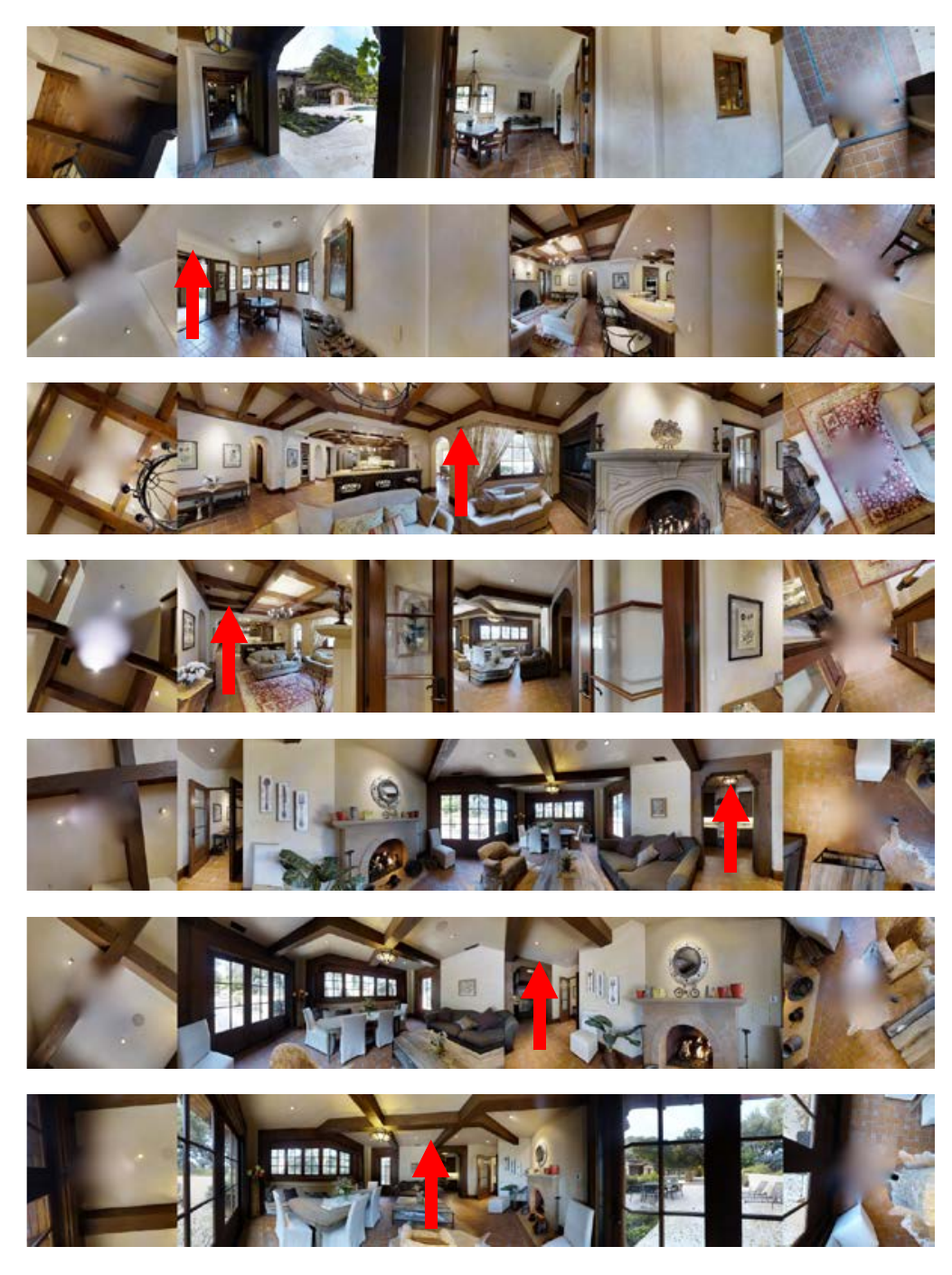}
         \caption{ground-truth trajectory}
     \end{subfigure}
     \hfill
     \vspace{0.2cm}
     \begin{subfigure}[b]{0.49\textwidth}
         \centering
         \includegraphics[width=\textwidth]{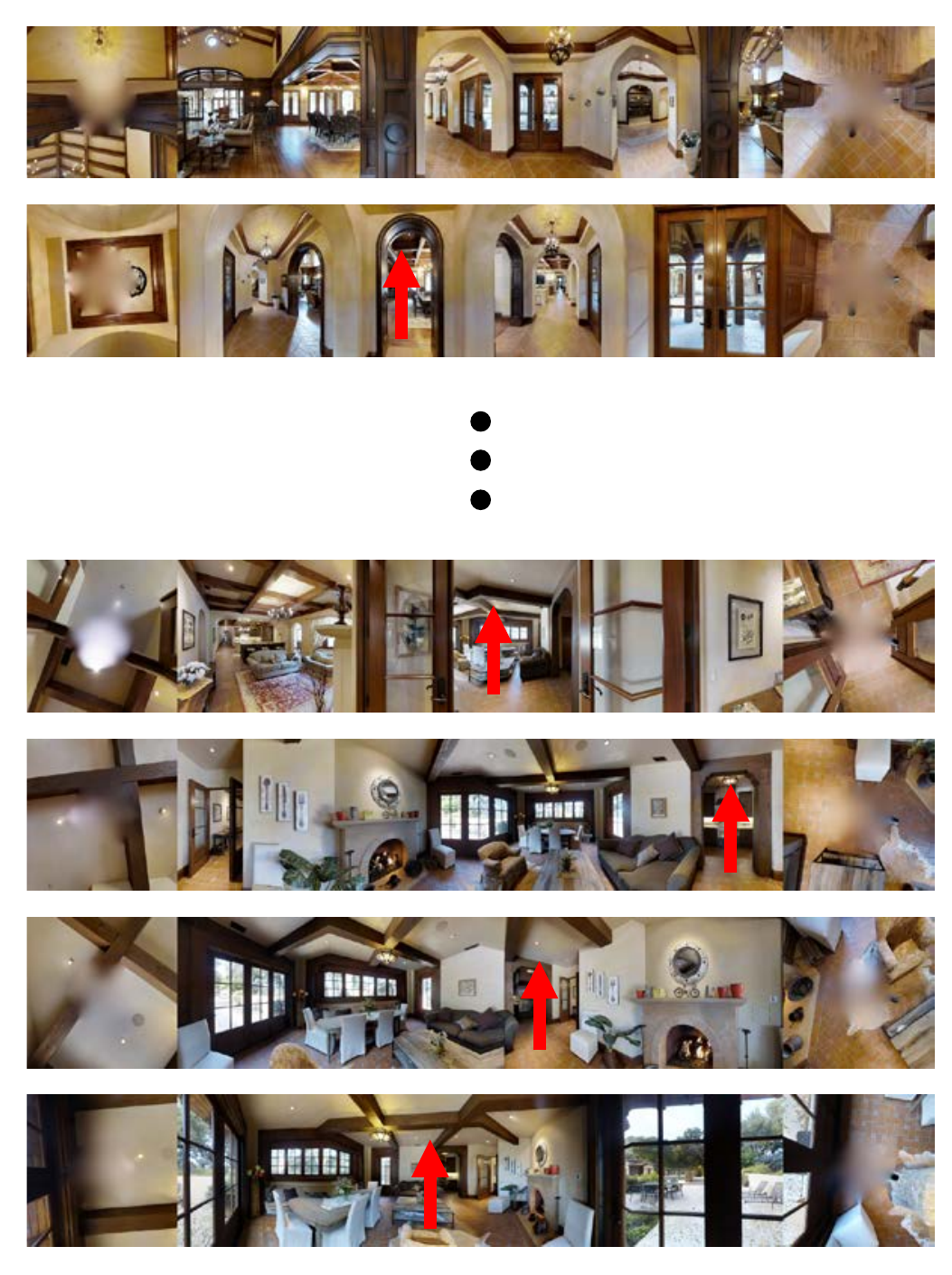}
         \caption{trajectory predicted by ESceme}
     \end{subfigure}
     \caption{Failure case in R2R val unseen split. The instruction is ``Leave sitting room and head towards the kitchen, turn right at living room and enter. Walk through living room to dining room and enter. Turn left and head to front door. Exit the house and stop on porch.'' After correctly predicting the first three actions, ESceme failed to enter the \textit{dining room} and got lost.}
     \label{fig:failure_r2r_1}
\end{figure*}

\bibliographystyle{abbrv}
\bibliography{main}
\end{document}